%% file: 0598.tex
\documentclass[runningheads]{llncs}
\usepackage{graphicx}
\usepackage{comment}
\usepackage{amsmath,amssymb} 
\usepackage{color}
\usepackage{hyperref}
\usepackage{comment}
\usepackage{booktabs}
\usepackage{boldline, multirow}
\usepackage{arydshln}

\newcommand\revision[1]{\color{black}#1}

\begin{document}
\pagestyle{headings}
\mainmatter
\def\ECCVSubNumber{598}  

\title{GMNet: Graph Matching Network for Large Scale Part Semantic Segmentation in the Wild} 

\titlerunning{GMNet for Large Scale Part Semantic Segmentation in the Wild}
%
\author{Umberto Michieli\orcidID{0000-0003-2666-4342} \and
Edoardo Borsato \and
Luca Rossi \and 
Pietro Zanuttigh\orcidID{0000-0002-9502-2389}}
\authorrunning{U. Michieli et al.}
%
\institute{Department of Information Engineering, University of Padova, Padova, Italy
\email{\{michieli, borsatoedo, rossiluc, zanuttigh\}@dei.unipd.it}}
\maketitle

\begin{abstract} 
The semantic segmentation of parts of objects in the wild is a challenging task in which multiple instances of objects and multiple parts within those objects must be detected in the scene. This problem remains nowadays very marginally explored, despite its fundamental importance towards detailed object understanding. In this work, we propose a novel framework combining higher object-level context conditioning and part-level spatial relationships to address the task. To tackle object-level ambiguity, a class-conditioning module is introduced to retain class-level semantics when learning parts-level semantics. In this way, mid-level features carry also this information prior to the decoding stage. To tackle part-level ambiguity and localization we propose a novel adjacency graph-based module that aims at matching the relative spatial relationships between ground truth and predicted parts. The experimental evaluation on the Pascal-Part dataset shows that we achieve state-of-the-art results on this task. 
\end{abstract}


\input{sections/introduction.tex}
\input{sections/related.tex}

\input{sections/method.tex}
\input{sections/graph.tex}
\input{sections/training.tex}
\input{sections/results_58.tex}

\input{sections/results_108.tex}

\input{sections/ablation.tex}

\input{sections/conclusion.tex}

\clearpage

%
%
\bibliographystyle{splncs04}
\bibliography{refs}

\pagebreak
\pagebreak

\onecolumn
\begin{center}
  \textbf{\large GMNet: Graph Matching Network for Large Scale Part Semantic Segmentation in the Wild \\ \vspace{0.1cm} \textit{Supplementary Material}}\\[.2cm]
  Umberto Michieli, Edoardo Borsato, Luca Rossi and Pietro Zanuttigh\\[.1cm]
  {\itshape University of Padova,\\ Department of Information Engineering, Via G. Gradenigo 6/b, 35131, Padova, Italy}
\end{center}

\setcounter{equation}{0}
\setcounter{figure}{0}
\setcounter{table}{0}
\setcounter{page}{1}
\renewcommand{\theequation}{S\arabic{equation}}
\renewcommand{\thefigure}{S\arabic{figure}}
\renewcommand{\thetable}{S\arabic{table}}

In this document we show some further experimental results. In particular we  report the Intersection over Union (IoU) and Pixel Accuracy (PA) for each  per-part-class and some averaged metrics as mean IoU (mIoU), mean PA (mPA) and mean Class Accuracy (mCA). The results are reported for both the Pascal-Part-58 and the Pascal-Part-108 datasets. Finally, some additional visual results on both datasets are presented.

\input{sections/suppl_58.tex}

\input{sections/suppl_108.tex}

\end{document}

%% file: sections/introduction.tex
\section{Introduction}
\label{sec:introduction}

Semantic segmentation is a wide research field and a huge number of approaches have been proposed for this task \cite{chen2018deeplab,zhao2017pyramid,long2015fully}. The  segmentation and labeling of parts of objects can be regarded as a special case of semantic segmentation that focuses on parts decomposition. The information about parts provides a richer representation 
for many fine-grained tasks, such as pose estimation \cite{dong2014towards,yang2011articulated}, category detection \cite{chen2014detect,azizpour2012object,zhang2014part}, fine-grained action detection \cite{wang2012discriminative} and image classification \cite{sun2013learning,krause2015fine}.
However, current approaches for semantic segmentation are not optimized to distinguish between different semantic parts since corresponding parts in different semantic classes often share similar appearance. Additionally, they only capture limited local context while the precise localization of semantic part layouts and their interactions {\revision requires} a wider perspective of the image. Thus, {\revision it is not sufficient to} take standard semantic segmentation methods and treat each part as an independent class.
In the literature, object-level semantic segmentation 
has been extensively studied. Part parsing, instead, has only been marginally explored in the context of a few specific single-class objects, such as humans \cite{liang2015deep,yamaguchi2012parsing,zhu2011max,eslami2012generative}, cars \cite{song2017embedding,lu2014parsing} and animals \cite{wang2015semantic,wang2015joint,haggag2016semantic}. Multi-class part-based semantic segmentation has only been considered in a recent work \cite{zhao2019multi}, due to the challenging scenario of part-level as well as object-level ambiguities. Here, we introduce an approach dealing with the semantic segmentation of  an even larger set of parts and we demonstrate that the proposed methodology is able to deal with a large amount of parts contained in the scenes.
 
Nowadays, one of the most active research directions is the transfer of previous knowledge, acquired on a different but related task, to a new situation. Different interpretations may exist to this regard. In the class-incremental task, the learned model is updated to perform a new task whilst preserving previous {\revision capabilities}: many methods have been proposed for image classification \cite{dhar2019learning,rebuffi2017icarl,li2018learning}, object detection \cite{shmelkov2017incremental} and semantic segmentation \cite{michieli2019incremental,michieli2020knowledge}. Another aspect regards the coarse-to-fine refinement at the semantic level, in which previous knowledge acquired on a coarser task is exploited to perform a finer task \cite{hariharan2015hypercolumns,xia2017joint,mel2020incremental}. In this paper, instead, we investigate the coarse-to-fine refinement at the spatial level, in which object-level classes are split into their respective parts \cite{wang2015joint,xia2016zoom,zhao2019multi}. 

More precisely, we  investigate the multi-object and multi-part parsing in the wild, which simultaneously handles all semantic objects and parts within each object in the scene. Even strong recent baselines for semantic segmentation, such as FCN \cite{long2015fully}, SegNet \cite{badrinarayanan2017segnet}, PSPNet \cite{zhao2017pyramid} or Deeplab \cite{chen2018deeplab,chen2017rethinking}, face additional challenges when dealing with this task, as shown in \cite{zhao2019multi}. In particular, the simultaneous appearance of multiple objects and the inter-class ambiguity may cause inaccurate boundary localization and severe classification errors. For instance, animals often have homogeneous appearance due to furs on the whole body. Additionally, the appearance of some parts over multiple object classes may be very similar, such as cow legs and sheep legs.
Current algorithms heavily suffer from these aspects. To address object-level ambiguity, we propose an object-level conditioning to serve as guidance for part parsing within the object. An auxiliary reconstruction module from parts to objects further penalize predictions of parts in regions occupied by an object which does not contain the predicted parts within it.
At the same time, to tackle part-level ambiguity, we introduce a graph-matching module to preserve the relative spatial relationships between ground truth and predicted parts.

When people look at scenes, they tend to locate first the objects and then to refine them via semantic part parsing \cite{xia2016zoom}. This is the same rationale for our class-conditioning approach, which consists of an approach to refine parts localization exploiting previous knowledge. In particular, the object-level predictions of the model serve as a conditioning term for the decoding stage on the part-level. The predictions are processed via an object-level semantic embedding Convolutional Neural Network (CNN) and its features are concatenated with the ones produced by the encoder of the part-level segmentation network. The extracted features are enriched with this type of information prior, guiding the output of the part-level decoding stage.
We further propose to address part-level ambiguity 
via a novel graph-matching technique applied to the segmentation maps. A couple of adjacency graphs are built from neighboring 
parts both from the ground-truth and from the predicted segmentation maps.
Such graphs are weighted with the normalized number of adjacent pixels to represent the strength of connection between the parts. Then, a novel loss function is designed to enforce their similarity.
These provisions allow the architecture to discover the differences in appearance between different parts within a single object, and at the same time to avoid the ambiguity across similar object categories.

The main contributions of this paper can be summarized as follows:
\begin{itemize}
\item We tackle the challenging multi-class part parsing via an object-level semantic embedding network conditioning the part-level decoding stage.

\item We introduce a novel graph-matching module to guide the learning process toward accurate relative localization of semantic parts.

\item Our approach (GMNet) achieves new state-of-the-art performance on  multi-object part parsing on the Pascal-Part dataset 
 \cite{chen2014detect}. Moreover, it scales well to large sets of parts. 
\end{itemize}

%% file: sections/related.tex
\section{Related Work}
\label{sec:related}

\textbf{Semantic Segmentation} is one of the key tasks for automatic scene understanding. Current techniques are based on the Fully Convolutional Network (FCN) framework \cite{long2015fully}, which firstly enabled accurate and end-to-end semantic segmentation. Recent works based on FCN, such as SegNet \cite{badrinarayanan2017segnet}, PSPNet \cite{zhao2017pyramid} and Deeplab \cite{chen2017rethinking,chen2018deeplab}, are typically regarded as the state-of-the-art architectures for semantic segmentation. Some recent reviews on the topic are \cite{liu2019recent,guo2018review}. \\
\textbf{Single-Object Part Parsing} has been actively investigated in the recent literature. However, most previous work assumes images containing only the considered object, well-localized beforehand and with no occlusions. Single-object parts parsing has been applied to animals \cite{wang2015semantic}, cars \cite{eslami2012generative,lu2014parsing,song2017embedding} and humans parsing \cite{liang2015deep,yamaguchi2012parsing,zhu2011max,eslami2012generative}.
Traditional deep neural network architectures may also be applied to part parsing regarding each semantic part as a separate class label. However, such strategies suffer from the high similar appearance between parts and from large scale variations of objects and parts. 
Some coarse-to-fine strategies have been proposed to tackle this issue. Hariharan et al. \cite{hariharan2015hypercolumns} propose to sequentially perform object detection, object segmentation and part segmentation with different architectures. However, there are some limitations, in particular the complexity of the training and  the error propagation throughout the pipeline. An upgraded version of the framework has been presented in \cite{xia2016zoom}, where the same structure is employed for the three networks and an automatic adaptation to the size of the object is introduced. In \cite{wang2015joint} a two-channels FCN is employed to jointly infer object and part segmentation for animals. However, it uses only {\revision a} single-scale network not capturing small parts and a fully connected CRF is used as post-processing technique to explore the relationship between parts and body to perform the final prediction. 
In \cite{chen2016attention} an attention mechanism that learns to softly weight the multi-scale features at each pixel location is proposed.

Some approaches resort to structure-based methodologies, e.g. compositional, to model part relations \cite{wang2015semantic,wang2015joint,liang2018look,liang2017interpretable,liang2016semantic,fang2018weakly}. Wang et al. \cite{wang2015semantic} propose a model to learn a mixture of compositional models under various poses and viewpoints for certain animal classes. 
In \cite{liang2018look} a self-supervised structure-sensitive learning approach is proposed to constrain human pose structures into parsing results. In \cite{liang2016semantic,liang2017interpretable} graph LSTMs are employed to refine the parsing results of initial over-segmented superpixel maps. Pose estimation is also useful for part parsing task \cite{xia2017joint,nie2018mutual,fang2018weakly,liang2018look,zhao2017self}. In \cite{xia2017joint}, the authors refine the segmentation maps by supervised pose estimation. In \cite{nie2018mutual} a mutual learning model is built for pose estimation and part segmentation. In \cite{fang2018weakly}, the authors exploit anatomical similarity among humans to transfer the parsing results of a person to another person with similar pose. In \cite{zhao2017self} multi-scale features aggregation at each pixel is combined with a self-supervised joint loss to further improve the feature discriminative capacity. 
Other approaches utilize tree-based approach to hierarchically partition the parts \cite{lu2014parsing,xia2015pose}. Lu et al. \cite{lu2014parsing} propose a method based on tree-structured graphical models from different viewpoints combined with segment appearance consistency for part parsing. Xia et al. \cite{xia2015pose} firstly generate part segment proposals and then infer the best ensemble of parts-segment  through and-or graphs.


Even though single-object part parsing has been extensively studied so far, \textbf{Multi-Object and Multi-Part Parsing} has been considered only recently \cite{zhao2019multi}. In this setup, most previous techniques fail struggling with objects that were not previously well-localized, isolated and with no occlusions. Zhao et al. in \cite{zhao2019multi} tackle this task via a joint parsing framework with boundary and semantic awareness for enhanced part localization and object-level guidance. Part boundaries are detected at early stages of feature extraction and then used in an attention mechanism to emphasize the features along the boundaries at the decoding stage. An additional attention module is employed to perform channel selection and is supervised by a supplementary branch predicting the semantic object classes.

%% file: sections/method.tex
\section{Method}
\label{sec:method}

When we look at images, we often firstly locate the objects and then the more detailed task of semantic part parsing is addressed using mainly two priors: (1) object-level information and (2) relative spatial relationships among parts.
Following this rationale, the semantic parts parsing is supported by the information coming from an initial prediction of the coarse object-level set of classes and by a graph-matching strategy working at the parts-level. 

\begin{figure}[t]
\centering
\includegraphics[width=1\textwidth]{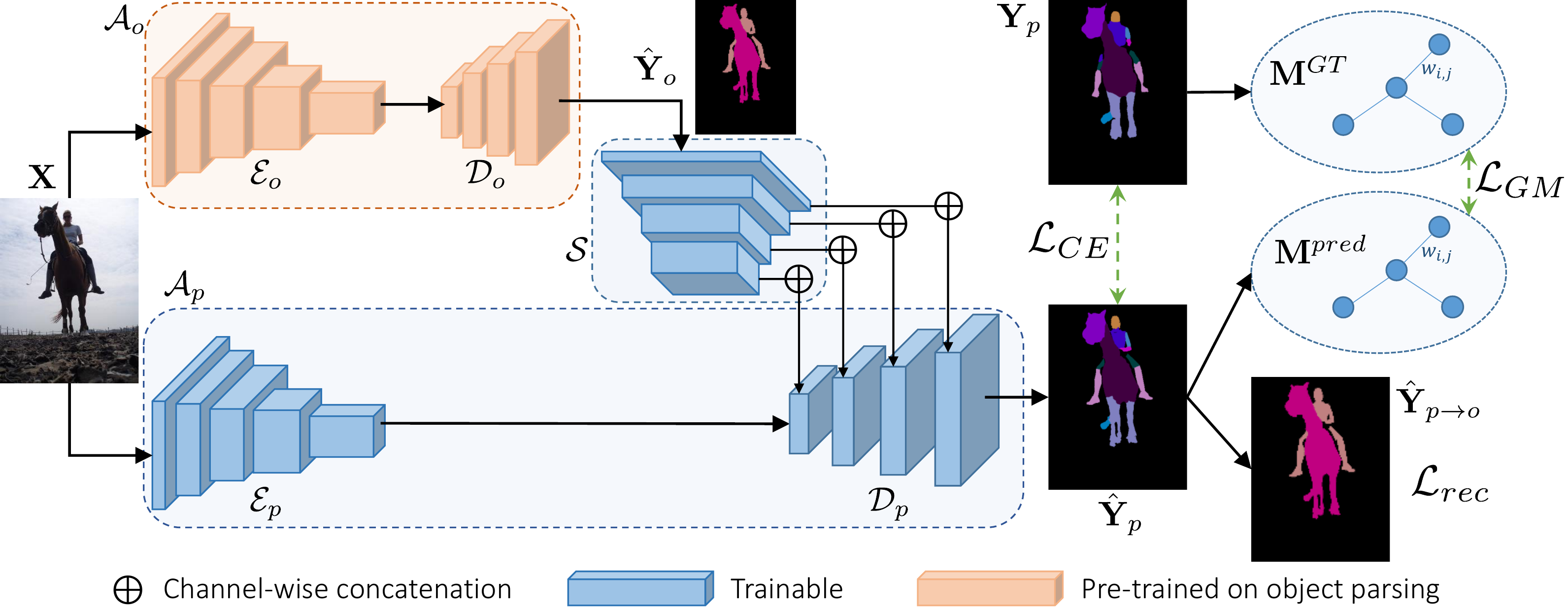}
\caption{Architecture of the proposed Graph Matching Network (GMNet) approach. A semantic embedding network takes as input the object-level segmentation map and acts as high level conditioning when learning the semantic segmentation of parts. On the right, a reconstruction loss function rearranges parts into objects and the graph matching module aligns the relative spatial relationships between ground truth and predicted parts.}
\label{fig:architecture}
\end{figure}

An overview of our framework is shown in Figure~\ref{fig:architecture}. We employ two semantic segmentation networks $\mathcal{A}_o$ and $\mathcal{A}_p$ trained for the objects-level and parts-level task respectively, together with a semantic embedding network $\mathcal{S}$ transferring and processing the information of the first network to the second to address the object-level prior. This novel coarse-to-fine strategy to gain insights into parts detection will be the subject of this section. Furthermore, we account for the second prior exploiting an adjacency graph structure to mimic the spatial relationship between neighboring parts to allow for a general overview of the semantic parts as described in Section~\ref{sec:graph}.

The semantic segmentation networks have an autoencoder structure and 
can be written as the composition of an encoder and a decoder as $\mathcal{A}_o=\{\mathcal{E}_o,\mathcal{D}_o\}$ and $\mathcal{A}_p=\{\mathcal{E}_p,\mathcal{D}_p\}$ for the object-level and part-level networks, respectively. We employ the Deeplab-v3 \cite{chen2017rethinking} {\revision segmentation network with}  Resnet-101 \cite{he2016deep} as encoder. The network $\mathcal{A}_o$ is trained 
using the object-level ground truth labels and then kept fixed. It extracts object-level segmentation maps which serve as a guidance for the decoder of the parts-level network $\mathcal{D}_p$, in order to avoid the ambiguity across similar object categories. 
We achieve this behavior by feeding the output maps of $\mathcal{A}_o$ to an object-level semantic embedding network. In this work, we used a CNN (denoted with $\mathcal{S}$)  formed by a cascade of $4$ convolutional layers with stride of $2$, square kernel sizes of $7$, $5$, $3$, $3$, and channel sizes of $128$, $256$, $512$, $1024$.

The parts-level semantic segmentation network $\mathcal{A}_p$ has the same encoder architecture of $\mathcal{A}_o$. Its decoder $\mathcal{D}_p$, instead, 
merges the features computed on the RGB image and the ones computed on the object-level predicted map via multiple channel-wise concatenations.
More in detail, each layer of the decoder considers a different resolution and its feature maps are concatenated with the layer at corresponding resolution of $\mathcal{S}$.
In this way, the combination is performed at multiple resolutions in the feature  space to achieve higher scale invariance as shown in Figure~\ref{fig:architecture}.  

Formally,
given an input RGB image $\mathbf{X}\in \mathbb{R}^{W\times H}$, the concatenation between part and object-level aware features is formulated as:
\begin{equation}
\mathcal{F}_i(\mathbf{X}) = \mathcal{D}_{p,i}(\mathbf{X}) \oplus \mathcal{S}_{k+1-i}(\mathcal{A}_o(\mathbf{X})) 
\quad \quad \quad
i=1,...,k
\end{equation}
where $\mathcal{D}_{p,i}$ is the $i$-th decoding layer of the part segmentation network, $\mathcal{S}_i$ denotes the $i$-th layer of $\mathcal{S}$, $k$ is the number of layers and matches the number of upsampling stages of the decoder (e.g., $k=4$ in the Deeplab-v3), $\mathcal{F}_i$ is the input of $\mathcal{D}_{p,i+1}$. 
Since the object-level segmentation is not perfect, in principle, errors from the predicted class in the object segmentation may propagate to the parts. To account for this, similarly to \cite{xia2016zoom}, here we do not make premature decisions but the channel-wise concatenation still leaves the final decision of the labeling task to the decoder. 

The training of the proposed framework (i.e., of $\mathcal{A}_p$ and $\mathcal{S}$, while $\mathcal{A}_o$ is kept fixed after the initial training)  is driven by multiple loss components. The first is a standard cross-entropy loss $\mathcal{L}_{CE}$ to learn the semantic segmentation of parts:
\begin{equation}
\label{eq:CE}
\mathcal{L}_{CE} = 
\sum^{N_p}_{ c_p = 1 }
\mathbf{Y}_p [c_p] \cdot 
\log \left( \hat{\mathbf{Y}}_p [c_p]   \right)
\end{equation}
where $\mathbf{Y}_p$ is the one-hot encoded ground truth map, $\hat{\mathbf{Y}}_p$ is the predicted map, $c_p$ is the part-class index and $N_p$ is the number of parts.

The object-level semantic embedding network is further guided by a reconstruction module that rearranges parts into objects. This is done applying a cross-entropy loss between object-level one-hot encoded ground truth maps $\mathbf{Y}_o$ and the summed probability $\hat{\mathbf{Y}}_{p\rightarrow o}$ derived from the part-level prediction. More formally, defining $l$ as the parts-to-objects mapping such that object $j$ contains parts from index $l[j-1]+1$ to $l[j]$, we can write the summed probability as:
\begin{equation}
\label{eq:summing_softmax}
\hat{\mathbf{Y}}_{p\rightarrow o}[j]    =     \sum_{i=l[j-1]+1,...,l[j]}       \hat{\mathbf{Y}}_p [i]  \quad\quad \quad j=1,...,N_o
\end{equation}
where $N_o$ is the number of object-level classes and $l[0]=0$. Then, we define the reconstruction loss as:
\begin{equation}
\label{eq:rec}
\mathcal{L}_{rec} = 
\sum^{N_o}_{ c_o = 1 }
\mathbf{Y}_o [c_o] \cdot 
\log \left( \hat{\mathbf{Y}}_{p\rightarrow o} [c_o]   \right)
\end{equation}
The auxiliary reconstruction function $\mathcal{L}_{rec}$ acts differently from the usual cross-entropy loss on the parts $\mathcal{L}_{CE}$. While $\mathcal{L}_{CE}$ penalizes wrong predictions of parts in all the portions of the image, $\mathcal{L}_{rec}$ only penalizes for part-level predictions located outside the respective object-level class. In other words, the event of predicting parts outside the respective object-level class is penalized by both the losses. Instead, parts predicted within the object class are penalized only by $\mathcal{L}_{CE}$, i.e., they are considered as a less severe type of error since, in this case, parts only {\revision need} to be properly localized inside the object-level class.

%% file: sections/graph.tex
\section{Graph-Matching for Semantic Parts Localization}
\label{sec:graph}

Providing global context information and disentangling relationships is useful to distinguish fine-grained parts. For instance, upper and lower arms share highly similar appearance. To differentiate between them, global and reciprocal information, like the relationship with neighboring parts, provides an effective context prior.
Hence, to further enhance the accuracy of part parsing, we tackle part-level ambiguity and localization by proposing a novel module based on an adjacency graph that matches the parts spatial relationships between ground truth and predicted parts. More in detail, the graphs capture the adjacency relationships between each couple of parts and then we enforce the matching between the ground truth and predicted graph through an additional loss term. Although graph matching is a very well studied problem \cite{emmert2016fifty,livi2013graph}, it has never been applied to this context before, i.e. as a loss function to drive deep learning architectures for semantic segmentation. The only other attempt to design a graph matching loss is \cite{das2018unsupervised}, which however  deals with a completely different task, i.e.,  domain adaptation in classification, and has a different interpretation of the graph, that 
measures the matching between the source and target domains.

\begin{figure}[t]
\centering
\includegraphics[width=1\textwidth]{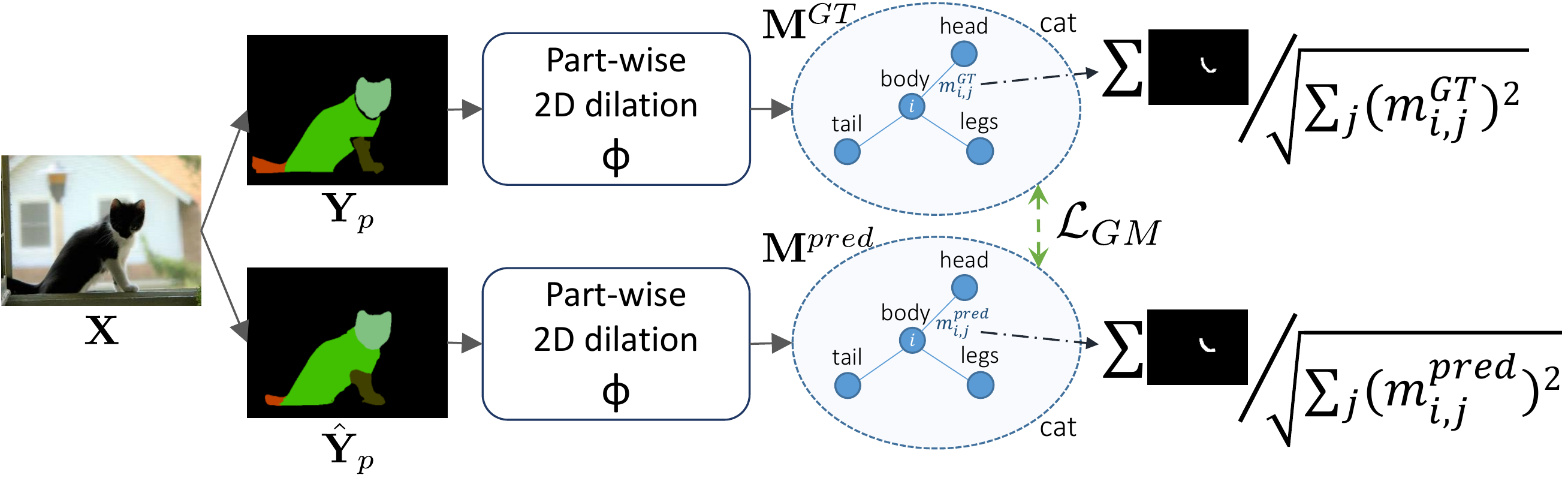}
\caption{Overview of the graph matching module. In this case, cat's head and body would be considered as detached without the proper morphological dilation over the parts.}
\label{fig:graph}
\end{figure}

An overview of this module is presented in Figure~\ref{fig:graph}. 
Formally, we represent the graphs using two (square) weighted adjacency matrices of size $N_p$:
\begin{equation}
\tilde{\mathbf{M}}^{GT}=\left\lbrace \tilde{m}^{GT}_{i,j} \right\rbrace_{\substack{i=1,...,N_p\\j=1,...,N_p}} \quad \quad \quad \ \ 
\tilde{\mathbf{M}}^{pred}=\left\lbrace \tilde{m}^{pred}_{i,j} \right\rbrace_{\substack{i=1,...,N_p\\j=1,...,N_p}}
\end{equation}
The first matrix ($\tilde{\mathbf{M}}^{GT}$) contains the adjacency information computed on ground truth data, while the second ($\tilde{\mathbf{M}}^{pred}$) has the same information computed on the predicted segmentation maps. 
Each element of the matrices provide a measure of how close the two parts $p_i$ and $p_j$ are in the ground truth and in the predicted segmentation maps, respectively.
We do not consider self-connections, hence $\tilde{m}^{GT}_{i,i}= \tilde{m}^{pred}_{i,i}=0$ for $i=1,...,N_p$.
To measure the closeness between couples of parts, that is {\revision a} hint of the strength of connection between them, we consider weighted matrices where each entry $\tilde{m}_{i,j}$ depends on the length of the contour in common between them. 
Actually, to cope for some inaccuracies inside the dataset where some adjacent parts are separated by thin background regions, the entries of the matrices are the counts of pixels belonging to one part with a distance less or equal than $T$ from a sample belonging to the other part. In other words, $\tilde{m}^{GT}_{i,j}$ represents the number of pixels in $p_i$ whose distance from a pixel in $p_j$ is less than $T$. We empirically set $T=4$ pixels. Since the matrix $\tilde{\mathbf{M}}^{pred}$  needs to be recomputed at each training step, we approximate this operation by dilating the two masks of $\lceil{T/2}\rceil$ and computing the intersecting region. Formally, defining with $p^{GT}_i = \mathbf{Y}_p[i]$ the mask of the $i$-th part in the ground truth map $\mathbf{Y}_p$, we have:
\begin{equation}
\tilde{m}^{GT}_{i,j}=\left|  \left\lbrace    
s\in \Phi\left(p_i^{GT}\right) \cap \Phi\left(p_j^{GT}\right)
\right\rbrace \right|
\end{equation} 

Where $s$ is a generic pixel, $\Phi(\cdot)$ is the morphological 2D dilation operation and $|\cdot|$ is the cardinality of the given set. We apply a row-wise L2 normalization and we obtain a matrix of \textit{proximity ratios} $\displaystyle \mathbf{M}^{GT}_{[i,:]} = \tilde{\mathbf{M}}^{GT}_{[i,:]}   /  \| \tilde{\mathbf{M}}^{GT}_{[i,:]} \|_2  $ that measures the flow from the considered part to all the others.

With this definition, non-adjacent parts have $0$ as entry.
The same approach is used for the adjacency matrix computed on the predicted segmentation map $\mathbf{M}^{pred}$ by  substituting $p^{GT}_i$ with $p^{pred}_i = \hat{\mathbf{Y}}_p [i]$.

Then, we simply define the Graph-Matching loss as the Frobenius norm between the two adjacency matrices:
\begin{equation}
\mathcal{L}_{GM} = \lvert\lvert \mathbf{M}^{GT} - \mathbf{M}^{pred} \rvert\rvert_F
\end{equation}
The aim of this loss function is to faithfully maintain the reciprocal relationships between parts. On  one hand, disjoint parts are enforced to be predicted as disjoint; on the other hand, neighboring parts are enforced to be predicted as neighboring and to match the proximity ratios.


{\revision Summarizing}, the overall training objective of our framework is:
\begin{equation}
\mathcal{L} = \mathcal{L}_{CE} + \lambda_1 \mathcal{L}_{rec}  + \lambda_2 \mathcal{L}_{GM} 
\end{equation}

where the hyper-parameters $\lambda_1$ and $\lambda_2$ are  used to control the relative contribution of the three losses to the overall objective function.

%% file: sections/training.tex
\section{Training of the Deep Learning Architecture}
\label{sec:training}

\subsection{Multi-Part Dataset}

For the experimental evaluation of the proposed multi-class part parsing framework we employed the Pascal-Part \cite{chen2014detect} dataset, which is {\revision currently} the largest dataset for this purpose. It contains a total of $10103$ variable-sized images with pixel-level parts annotation on the $20$ Pascal VOC2010 \cite{pascalvoc2010} semantic object classes (plus the \textit{background} class). We employ the original split from  \cite{chen2014detect}  with $4998$ images in the \textit{trainset} for training and  $5105$ images in the \textit{valset} for testing.
We consider two different sets of labels for this dataset. {\revision Firstly,} following \cite{zhao2019multi},
which is the only work dealing with the multi-class part parsing problem, we grouped the original semantic classes into $58$ part classes in total. Additionally, to further test our {\revision method} on a even more challenging scenario, we consider the grouping rules proposed by \cite{gonzalez2018semantic} for part detection that, instead, leads to a larger set of $108$ parts. 

\subsection{Training Details}

The modules introduced in this work are agnostic to the underlying network architecture and can be extended to other scenarios. For fair comparison with \cite{zhao2019multi} we employ a Deeplab-v3 \cite{chen2017rethinking} architecture with ResNet101 \cite{he2016deep} as the backbone. We follow the same training schemes of \cite{chen2018deeplab,chen2017rethinking,zhao2019multi} and we started from the official TensorFlow \cite{abadi2016tensorflow} implementation of the Deeplab-v3 \cite{chen2017rethinking,deeplab_research}. The ResNet101 was pre-trained on ImageNet \cite{imagenet2009} and its weights are available at \cite{deeplab_research}. During training, images are randomly left-right flipped and scaled of a factor from $0.5$ to $2$ times the original resolution with bilinear interpolation. The results in the testing stage are reported at the original image resolution. The model is trained for $50K$ steps with the base learning rate set to $5\cdot 10^{-3}$ and decreased with a polynomial decay rule with power $0.9$. We employ weight decay regularization of $10^{-4}$.
The atrous rate in the Atrous Spatial Pyramid Poooling (ASPP) is set to $(6,12,18)$ as in \cite{chen2018deeplab,zhao2019multi}.
We use a batch size of $10$ images and we set $\lambda_1=10^{-3}$ and $\lambda_2 = 10^{-1}$ to balance part segmentation. 
For the evaluation metric, we employ the mean Intersection over Union (mIoU) since pixel accuracy is dominated by large regions and little sensitive to the segmentation on many small parts, that are instead the main target of this work. 
The code and {\revision the part labels are} publicly available at \url{https://lttm.dei.unipd.it/paper_data/GMNet}.

%% file: sections/results_58.tex
\section{Experimental Results}
\label{sec:results}

In this section we show the experimental results on the multi-class part parsing task in two different scenarios with $58$ and $108$ parts respectively. We also present some ablation studies to verify the effectiveness of the proposed methodologies.

\begin{table*}[b]
\caption{IoU results on the Pascal-Part-58 benchmark. mIoU: mean per-part-class IoU. Avg: average per-object-class mIoU.} 
\label{tab:Pascal_part_58}
\setlength{\tabcolsep}{1.5pt}
\centering
\tiny
\renewcommand{\arraystretch}{1.5}
\begin{tabular}{|l|ccccccccccccccccccccc|cc|}
\hline
Method & \rotatebox{90}{bgr} &  \rotatebox{90}{aero} &  \rotatebox{90}{bike} &  \rotatebox{90}{bird} &\rotatebox{90}{boat} & \rotatebox{90}{bottle} & \rotatebox{90}{bus} 
  &\rotatebox{90}{car} & \rotatebox{90}{cat} & \rotatebox{90}{chair} & \rotatebox{90}{cow} & \rotatebox{90}{d. table}& \rotatebox{90}{dog} & \rotatebox{90}{horse} 
  & \rotatebox{90}{mbike} & \rotatebox{90}{person} & \rotatebox{90}{plant} &  \rotatebox{90}{sheep} & \rotatebox{90}{sofa} & \rotatebox{90}{train} & \rotatebox{90}{tv} & \rotatebox{90}{\textbf{mIoU}} & \rotatebox{90}{\textbf{Avg.}}\\
 \hline

SegNet\cite{badrinarayanan2017segnet} & \fontfamily{bch}\selectfont 85.4 & \fontfamily{bch}\selectfont 13.7 & \fontfamily{bch}\selectfont 40.7 & \fontfamily{bch}\selectfont 11.3 & \fontfamily{bch}\selectfont 21.7 & \fontfamily{bch}\selectfont 10.7 & \fontfamily{bch}\selectfont 36.7 & \fontfamily{bch}\selectfont 26.3 & \fontfamily{bch}\selectfont 28.5 & \fontfamily{bch}\selectfont 16.6 & \fontfamily{bch}\selectfont 8.9 & \fontfamily{bch}\selectfont 16.6 & \fontfamily{bch}\selectfont 24.2 & \fontfamily{bch}\selectfont 18.8 & \fontfamily{bch}\selectfont 44.7 & \fontfamily{bch}\selectfont 35.4 & \fontfamily{bch}\selectfont 16.1 & \fontfamily{bch}\selectfont 17.3 & \fontfamily{bch}\selectfont 15.7 & \fontfamily{bch}\selectfont 41.3 & \fontfamily{bch}\selectfont 26.1 & \fontfamily{bch}\selectfont 24.4 & \fontfamily{bch}\selectfont 26.5 \\

FCN\cite{long2015fully} & \fontfamily{bch}\selectfont 87.0 & \fontfamily{bch}\selectfont 33.9 & \fontfamily{bch}\selectfont 51.5 & \fontfamily{bch}\selectfont 37.7 & \fontfamily{bch}\selectfont 47.0 & \fontfamily{bch}\selectfont 45.3 & \fontfamily{bch}\selectfont 50.8 & \fontfamily{bch}\selectfont 39.1 & \fontfamily{bch}\selectfont 45.2 & \fontfamily{bch}\selectfont 29.4 & \fontfamily{bch}\selectfont 31.2 & \fontfamily{bch}\selectfont 32.5 & \fontfamily{bch}\selectfont 42.4 & \fontfamily{bch}\selectfont 42.2 & \fontfamily{bch}\selectfont 58.2 & \fontfamily{bch}\selectfont 40.3 & \fontfamily{bch}\selectfont 38.3 & \fontfamily{bch}\selectfont 43.4 & \fontfamily{bch}\selectfont 35.7 & \fontfamily{bch}\selectfont 66.7 & \fontfamily{bch}\selectfont 44.2 & \fontfamily{bch}\selectfont 42.3 & \fontfamily{bch}\selectfont 44.9 \\

DeepLab\cite{chen2018deeplab} & \fontfamily{bch}\selectfont 89.8 & \fontfamily{bch}\selectfont 40.7 & \fontfamily{bch}\selectfont 58.1 & \fontfamily{bch}\selectfont 43.8 & \fontfamily{bch}\selectfont 53.9 & \fontfamily{bch}\selectfont 44.5 & \fontfamily{bch}\selectfont 62.1 & \fontfamily{bch}\selectfont 45.1 & \fontfamily{bch}\selectfont 52.3 & \fontfamily{bch}\selectfont 36.6 & \fontfamily{bch}\selectfont 41.9 & \fontfamily{bch}\selectfont 38.7 & \fontfamily{bch}\selectfont 49.5 & \fontfamily{bch}\selectfont 53.9 & \fontfamily{bch}\selectfont 66.1 & \fontfamily{bch}\selectfont 49.0 & \fontfamily{bch}\selectfont 45.3 & \fontfamily{bch}\selectfont 45.3 & \fontfamily{bch}\selectfont 40.5 & \fontfamily{bch}\selectfont 76.8 & \fontfamily{bch}\selectfont 56.5 & \fontfamily{bch}\selectfont 49.9 & \fontfamily{bch}\selectfont 51.9 \\

BSANet\cite{zhao2019multi} & \fontfamily{bch}\selectfont 91.6 & \fontfamily{bch}\selectfont \textbf{50.0} & \fontfamily{bch}\selectfont 65.7 & \fontfamily{bch}\selectfont \textbf{54.8} & \fontfamily{bch}\selectfont 60.2 & \fontfamily{bch}\selectfont 49.2 & \fontfamily{bch}\selectfont 70.1 & \fontfamily{bch}\selectfont \textbf{53.5} & \fontfamily{bch}\selectfont \textbf{63.8} & \fontfamily{bch}\selectfont 36.5 & \fontfamily{bch}\selectfont 52.8 & \fontfamily{bch}\selectfont 43.7 & \fontfamily{bch}\selectfont 58.3* 
& \fontfamily{bch}\selectfont \textbf{66.0} & \fontfamily{bch}\selectfont 71.6* 
 & \fontfamily{bch}\selectfont \textbf{58.4} & \fontfamily{bch}\selectfont 55.0 & \fontfamily{bch}\selectfont 49.6 & \fontfamily{bch}\selectfont 43.1 & \fontfamily{bch}\selectfont 82.2 & \fontfamily{bch}\selectfont 61.4 & \fontfamily{bch}\selectfont 58.2 & \fontfamily{bch}\selectfont 58.9* \\\hline

Baseline\cite{chen2017rethinking} & \fontfamily{bch}\selectfont 91.1 & \fontfamily{bch}\selectfont 45.7 & \fontfamily{bch}\selectfont 63.2 &\fontfamily{bch}\selectfont  49.0 &\fontfamily{bch}\selectfont  54.4 &\fontfamily{bch}\selectfont  49.8 &\fontfamily{bch}\selectfont  67.6 &\fontfamily{bch}\selectfont  49.2 &\fontfamily{bch}\selectfont  59.8 &\fontfamily{bch}\selectfont  35.4 &\fontfamily{bch}\selectfont  47.6 &\fontfamily{bch}\selectfont  43.0 &\fontfamily{bch}\selectfont  54.4 &\fontfamily{bch}\selectfont  62.0 &\fontfamily{bch}\selectfont  68.0 &\fontfamily{bch}\selectfont  55.0 &\fontfamily{bch}\selectfont  48.9 &\fontfamily{bch}\selectfont  45.9 &\fontfamily{bch}\selectfont  43.2 &\fontfamily{bch}\selectfont  79.6 &\fontfamily{bch}\selectfont  57.7 &\fontfamily{bch}\selectfont  54.4 &\fontfamily{bch}\selectfont  55.7  \\

GMNet  & \fontfamily{bch}\selectfont \textbf{92.7} & \fontfamily{bch}\selectfont 46.7 & \fontfamily{bch}\selectfont \textbf{66.4} & \fontfamily{bch}\selectfont 52.0 & \fontfamily{bch}\selectfont	\textbf{70.0} & \fontfamily{bch}\selectfont \textbf{55.7} & \fontfamily{bch}\selectfont \textbf{71.1} & \fontfamily{bch}\selectfont 52.2 & \fontfamily{bch}\selectfont 63.2 & \fontfamily{bch}\selectfont \textbf{51.4} & \fontfamily{bch}\selectfont \textbf{54.8} & \fontfamily{bch}\selectfont \textbf{51.3} & \fontfamily{bch}\selectfont \textbf{59.6} & \fontfamily{bch}\selectfont 64.4 & \fontfamily{bch}\selectfont \textbf{73.9} & \fontfamily{bch}\selectfont 56.2 & \fontfamily{bch}\selectfont \textbf{56.2} & \fontfamily{bch}\selectfont \textbf{53.6} & \fontfamily{bch}\selectfont \textbf{56.1} & \fontfamily{bch}\selectfont \textbf{85.0} & \fontfamily{bch}\selectfont \textbf{65.6} & \fontfamily{bch}\selectfont \textbf{59.0} & \fontfamily{bch}\selectfont \textbf{61.8} \\
\hline
\end{tabular}
\begin{flushright}
*: values different from \cite{zhao2019multi} since they were wrongly reported in the paper. \hspace{0.1cm}
\end{flushright}
\end{table*}

\subsection{Pascal-Part-58}

To evaluate our framework we start from the scenario with 58 parts, i.e., the same experimental setting used in \cite{zhao2019multi}.
In Table~\ref{tab:Pascal_part_58} we compare the proposed model with existing semantic segmentation schemes. As evaluation criteria we employ the mean IoU of all the parts (i.e., mIoU), the average IoU for all the parts belonging to each single object, and the mean of these values (denoted as Avg., i.e., in this case each object has the same weight independently of the number of parts). Part-level metrics are reported in the supplementary material. 
As expected, traditional semantic segmentation architectures such as FCN \cite{long2015fully}, SegNet \cite{badrinarayanan2017segnet} and DeepLab \cite{chen2018deeplab} are not able to perform a fully satisfactory part-parsing. We adopt as our baseline network the DeepLab-v3 architecture \cite{chen2017rethinking}, that is the best performing among the compared standard approaches achieving $54.4\%$ of mIoU. 
The proposed GMNet approach combining both the object-level semantic embedding and the graph matching module achieves a higher accuracy of $59.0\%$ of mIoU, significantly outperforming all the other methods and in particular the baseline on every class with {\revision a} gap of $4.6\%$ of mIoU. 
The only other method {\revision specifically addressing} part-based semantic segmentation is BSANet \cite{zhao2019multi}, which achieves a lower mIoU of $58.2\%$.    {\revision Our method achieves} higher results over most of the objects, both with many parts  (like \textit{cow}, \textit{dog} and \textit{sheep}) and with no or few parts (like \textit{boat}, \textit{bottle}, \textit{chair}, \textit{dining table} and \textit{sofa}).

\newcommand{\sizefiggg}{0.19}
\begin{figure}[tbp]{}
\setlength\tabcolsep{1.5pt} 
\centering
\begin{tabular}{ccccc}
  RGB & Annotation & Baseline \cite{chen2017rethinking} & BSANet \cite{zhao2019multi} & GMNet (ours) \\

   \includegraphics[width=\sizefiggg\linewidth]{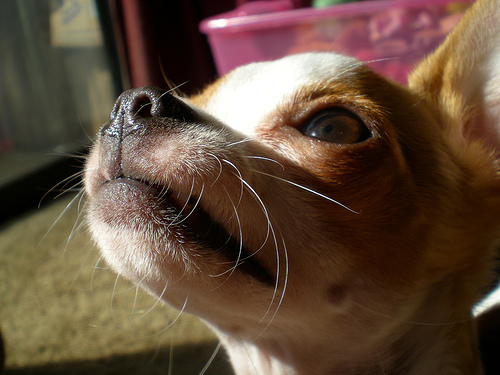} &
  \includegraphics[width=\sizefiggg\linewidth]{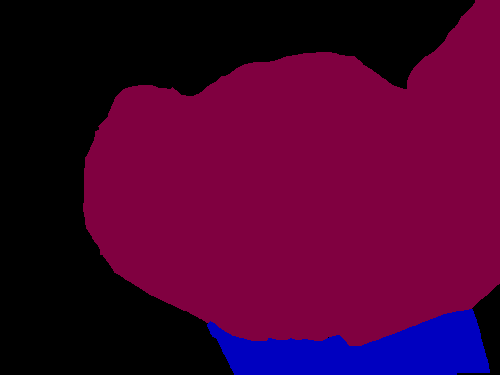} &
  \includegraphics[width=\sizefiggg\linewidth]{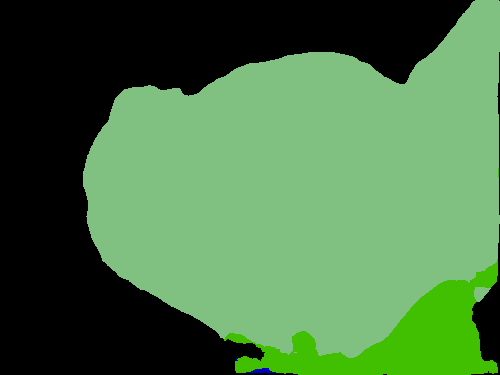} &
  \includegraphics[width=\sizefiggg\linewidth]{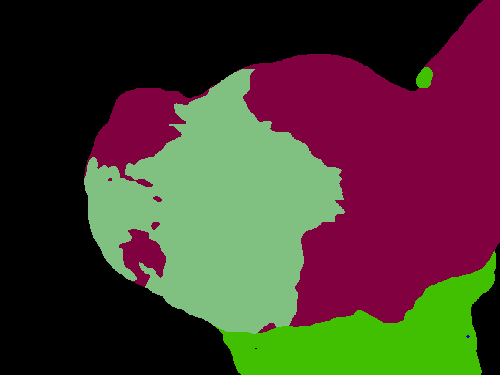} &
  \includegraphics[width=\sizefiggg\linewidth]{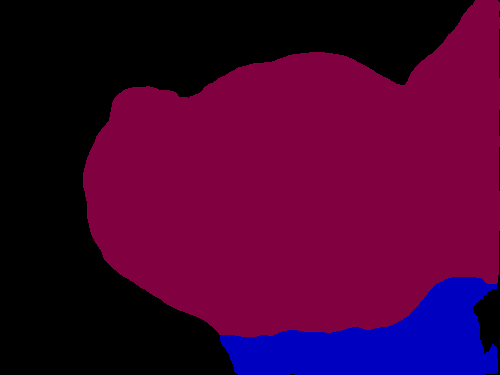} \\  
  
     \includegraphics[width=\sizefiggg\linewidth]{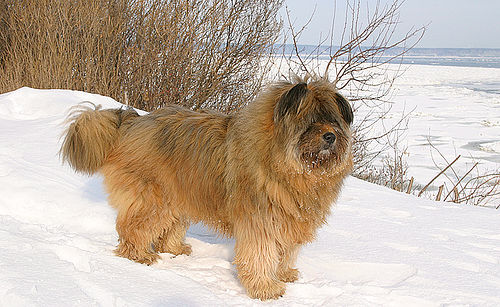} &
  \includegraphics[width=\sizefiggg\linewidth]{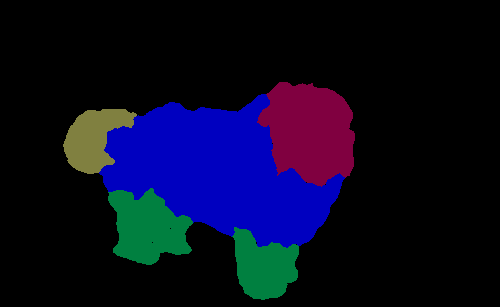} &
  \includegraphics[width=\sizefiggg\linewidth]{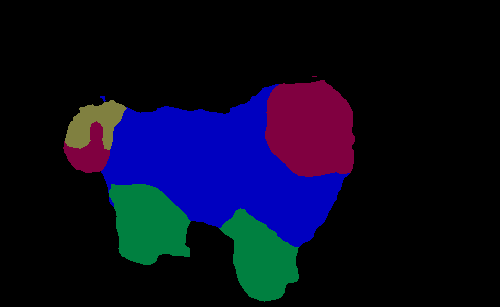} &
  \includegraphics[width=\sizefiggg\linewidth]{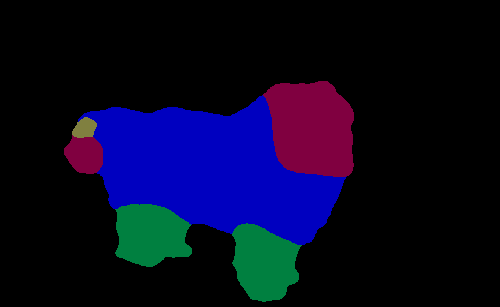} &
  \includegraphics[width=\sizefiggg\linewidth]{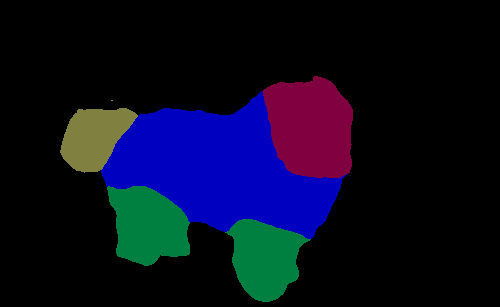} \\

  \includegraphics[width=\sizefiggg\linewidth]{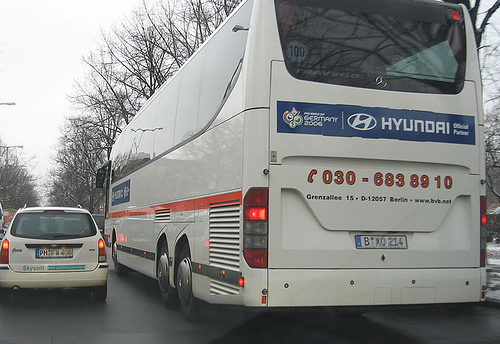} &
  \includegraphics[width=\sizefiggg\linewidth]{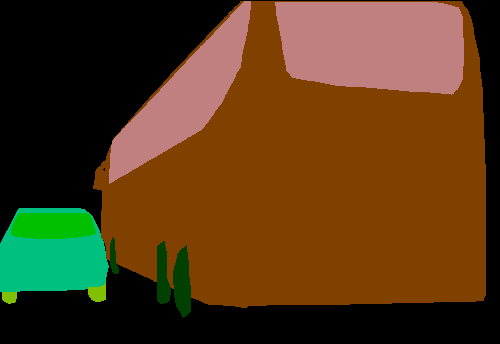} &
  \includegraphics[width=\sizefiggg\linewidth]{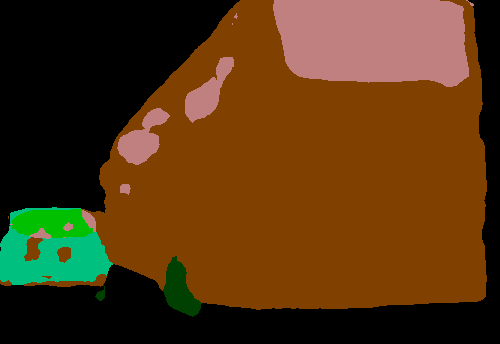} &
  \includegraphics[width=\sizefiggg\linewidth]{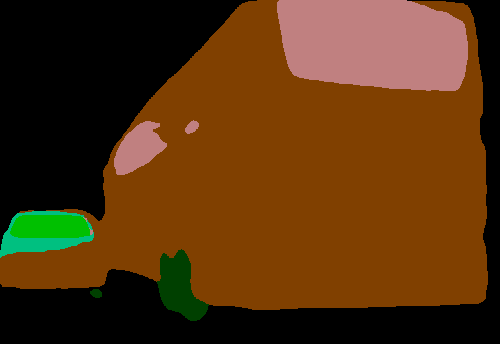} &
  \includegraphics[width=\sizefiggg\linewidth]{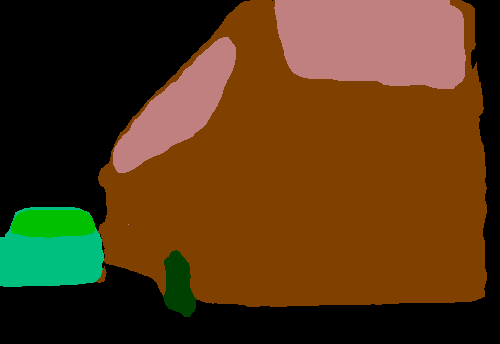} \\

     \includegraphics[width=\sizefiggg\linewidth]{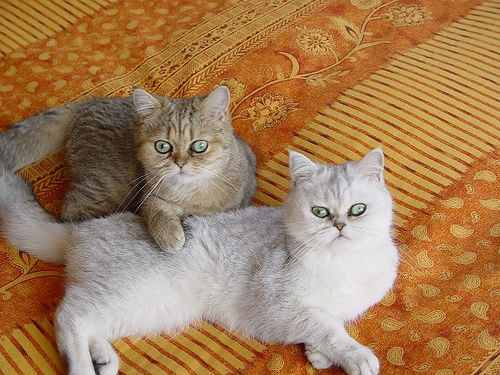} &
  \includegraphics[width=\sizefiggg\linewidth]{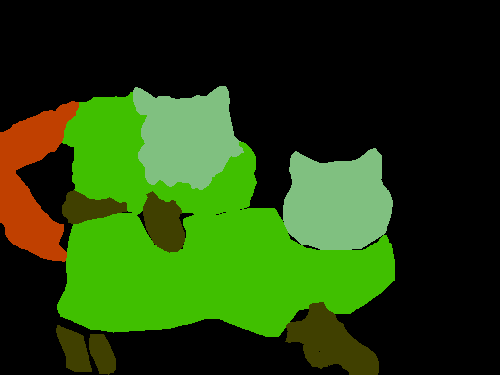} &
  \includegraphics[width=\sizefiggg\linewidth]{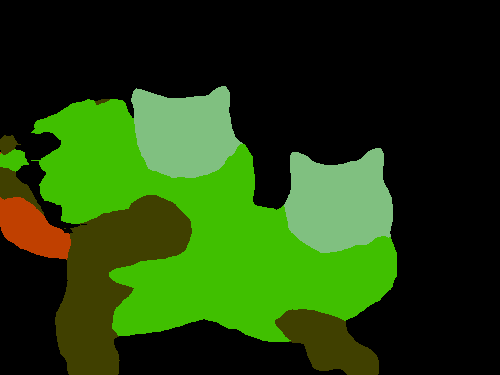} &
  \includegraphics[width=\sizefiggg\linewidth]{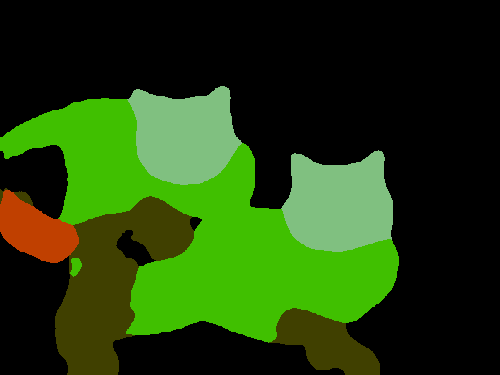} &
  \includegraphics[width=\sizefiggg\linewidth]{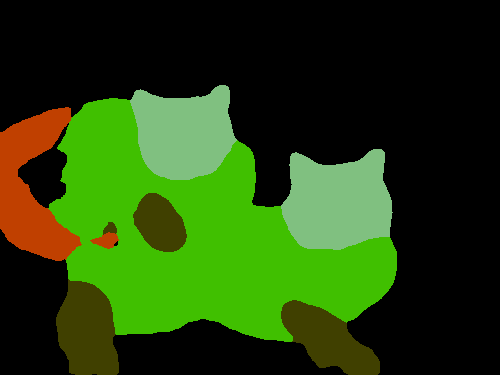} \\

 \end{tabular}
\caption{{\revision Segmentation results from} the Pascal-Part-58 dataset (\textit{best viewed in colors}).}
\label{fig:Pascal_part_58}
\end{figure}

Some qualitative results are shown in Figure~\ref{fig:Pascal_part_58} {\revision while additional ones are  in the supplementary material}. The figure allows to appreciate the effects of the two main {\revision contributions},  the semantic embedding and the graph matching modules. 

From one side, the object-level semantic embedding network brings useful additional information prior to the part-level decoding stage, thus enriching the extracted features to be object discriminative.
We can appreciate this aspect from the first and the third row. In the first row, the baseline completely misleads a dog with a cat (light green corresponds to \textit{cat\_head} and green to \textit{cat\_torso}). BSANet is able to partially recover the \textit{dog\_head} (amaranth corresponds to the proper labeling). Our method, instead, is able to accurately detect and segment the dog parts (\textit{dog\_head} in amaranth and \textit{dog\_torso} in blue) thanks to object-level priors coming from the semantic embedding module.
A similar discussion can be done also on the third image, where the baseline confuses car parts (green corresponds to \textit{window}, aquamarine to \textit{body} and light green to \textit{wheel}) with bus parts (pink is the \textit{window}, brown the \textit{body} and dark green the \textit{wheel}) and BSANet is not able to correct this error. GMNet, instead, can identify the correct object-level class and the respective parts, excluding the very small and challenging \textit{car\_wheels}, and at the same time can better segment the \textit{bus\_window}.

From the other side, the graph matching module helps in the mutual localization of parts within the same object-level class.
The effect of the graph matching module is more evident in the second and fourth row. In the second image, we can verify how both the baseline and BSANet are not able to correctly place the \textit{dog\_tail} (in yellow) misleading it with the \textit{dog\_head} (in red). Thanks to the graph matching module, GMNet can disambiguate between such parts and correctly exploit their spatial relationship with respect to the \textit{dog\_body}. 
In the fourth image, both the baseline and BSANet tend to overestimate the presence of the \textit{cat\_legs} (in dark green) and they miss one \textit{cat\_tail}. The constraints on the relative position among the various parts enforced by the graph matching module allow GMNet to properly segment and label the \textit{cat\_tail} and to partially correct the estimate of the \textit{cat\_legs}.

%% file: sections/results_108.tex
\subsection{Pascal-Part-108}

\begin{table*}[bp]
\caption{IoU results on the Pascal-Part-108 benchmark. mIoU: mean per-part-class IoU. Avg: average per-object-class mIoU. $\dag$: re-trained on the Pascal-Part-108 dataset.}
\label{tab:Pascal_part_108}
\setlength{\tabcolsep}{1.5pt}
\renewcommand{\arraystretch}{1.5}
\centering
\tiny
\begin{tabular}{|l|ccccccccccccccccccccc|cc|}
\hline
Method & \rotatebox{90}{bgr} &  \rotatebox{90}{aero} &  \rotatebox{90}{bike} &  \rotatebox{90}{bird} &\rotatebox{90}{boat} & \rotatebox{90}{bottle} & \rotatebox{90}{bus} 
  &\rotatebox{90}{car} & \rotatebox{90}{cat} & \rotatebox{90}{chair} & \rotatebox{90}{cow} & \rotatebox{90}{d. table}& \rotatebox{90}{dog} & \rotatebox{90}{horse} 
  & \rotatebox{90}{mbike} & \rotatebox{90}{person} & \rotatebox{90}{plant} &  \rotatebox{90}{sheep} & \rotatebox{90}{sofa} & \rotatebox{90}{train} & \rotatebox{90}{tv} & \rotatebox{90}{\textbf{mIoU}} & \rotatebox{90}{\textbf{Avg.}}\\
 \hline

\revision SegNet\cite{badrinarayanan2017segnet} & \fontfamily{bch}\selectfont 85.3 & \fontfamily{bch}\selectfont 11.2 & \fontfamily{bch}\selectfont 32.4 & \fontfamily{bch}\selectfont 6.3 & \fontfamily{bch}\selectfont 21.4 & \fontfamily{bch}\selectfont 10.3 & \fontfamily{bch}\selectfont 27.9 & \fontfamily{bch}\selectfont 22.6 & \fontfamily{bch}\selectfont 22.8 & \fontfamily{bch}\selectfont 17.0 & \fontfamily{bch}\selectfont 6.3 & \fontfamily{bch}\selectfont 12.5 & \fontfamily{bch}\selectfont 21.1 & \fontfamily{bch}\selectfont 14.9 & \fontfamily{bch}\selectfont 12.2 & \fontfamily{bch}\selectfont 32.2 & \fontfamily{bch}\selectfont 13.8 & \fontfamily{bch}\selectfont 12.6 & \fontfamily{bch}\selectfont 15.2 & \fontfamily{bch}\selectfont 11.3 & \fontfamily{bch}\selectfont 27.5 & \fontfamily{bch}\selectfont 18.6 & \fontfamily{bch}\selectfont 20.8 \\

\revision FCN\cite{long2015fully} & \fontfamily{bch}\selectfont 86.8 & \fontfamily{bch}\selectfont 30.3 & \fontfamily{bch}\selectfont 35.6 & \fontfamily{bch}\selectfont 23.6 & \fontfamily{bch}\selectfont 47.5 & \fontfamily{bch}\selectfont 44.5 & \fontfamily{bch}\selectfont 21.3 & \fontfamily{bch}\selectfont 34.5 & \fontfamily{bch}\selectfont 35.8 & \fontfamily{bch}\selectfont 26.6 & \fontfamily{bch}\selectfont 20.3 & \fontfamily{bch}\selectfont 24.4 & \fontfamily{bch}\selectfont 37.7 & \fontfamily{bch}\selectfont 29.8 & \fontfamily{bch}\selectfont 14.2 & \fontfamily{bch}\selectfont 35.6 & \fontfamily{bch}\selectfont 34.4 & \fontfamily{bch}\selectfont 28.9 & \fontfamily{bch}\selectfont 34.0 & \fontfamily{bch}\selectfont 18.1 & \fontfamily{bch}\selectfont 45.6 & \fontfamily{bch}\selectfont 31.6 & \fontfamily{bch}\selectfont 33.8 \\

\revision Deeplab\cite{chen2018deeplab} & \fontfamily{bch}\selectfont 90.2 & \fontfamily{bch}\selectfont 38.3 & \fontfamily{bch}\selectfont 35.4 & \fontfamily{bch}\selectfont 29.4 & \fontfamily{bch}\selectfont 57.0 & \fontfamily{bch}\selectfont 41.5 & \fontfamily{bch}\selectfont 27.0 & \fontfamily{bch}\selectfont 40.1 & \fontfamily{bch}\selectfont 45.5 & 
\fontfamily{bch}\selectfont 36.6 & \fontfamily{bch}\selectfont 33.3 & \fontfamily{bch}\selectfont 35.2 & \fontfamily{bch}\selectfont 41.1 & \fontfamily{bch}\selectfont 48.8 & \fontfamily{bch}\selectfont 19.5 & \fontfamily{bch}\selectfont 40.6 & \fontfamily{bch}\selectfont 46.0 & \fontfamily{bch}\selectfont 23.7 & \fontfamily{bch}\selectfont 40.8 & \fontfamily{bch}\selectfont 17.5 & \fontfamily{bch}\selectfont 70.0 & \fontfamily{bch}\selectfont 35.7 & \fontfamily{bch}\selectfont 40.8 \\

BSANet$\dag$ \cite{zhao2019multi} & \fontfamily{bch}\selectfont 91.6 & \fontfamily{bch}\selectfont 45.3 & \fontfamily{bch}\selectfont 40.9 & \fontfamily{bch}\selectfont \textbf{41.0} & \fontfamily{bch}\selectfont 61.4 & \fontfamily{bch}\selectfont 48.9 & \fontfamily{bch}\selectfont 32.2 & \fontfamily{bch}\selectfont 43.3 & \fontfamily{bch}\selectfont 50.7 & \fontfamily{bch}\selectfont 34.1 & \fontfamily{bch}\selectfont 39.4 & \fontfamily{bch}\selectfont 45.9 & \fontfamily{bch}\selectfont \textbf{52.1} & \fontfamily{bch}\selectfont 50.0 & \fontfamily{bch}\selectfont 23.1 & \fontfamily{bch}\selectfont 52.4 & \fontfamily{bch}\selectfont 50.6 & \fontfamily{bch}\selectfont 37.8 & \fontfamily{bch}\selectfont 44.5 & \fontfamily{bch}\selectfont 20.7 & \fontfamily{bch}\selectfont 66.3 & \fontfamily{bch}\selectfont 42.9 & \fontfamily{bch}\selectfont 46.3  \\\hline

Baseline \cite{chen2017rethinking} & \fontfamily{bch}\selectfont 90.9 & \fontfamily{bch}\selectfont 41.9 & \fontfamily{bch}\selectfont 44.5 & \fontfamily{bch}\selectfont 35.3 & \fontfamily{bch}\selectfont 53.7 & \fontfamily{bch}\selectfont 47.0 & \fontfamily{bch}\selectfont 34.1 & \fontfamily{bch}\selectfont 42.3 & \fontfamily{bch}\selectfont 49.2 & \fontfamily{bch}\selectfont 35.4 & \fontfamily{bch}\selectfont 39.8 & \fontfamily{bch}\selectfont 33.0 & \fontfamily{bch}\selectfont 48.2 & \fontfamily{bch}\selectfont 48.8 & \fontfamily{bch}\selectfont 23.2 & \fontfamily{bch}\selectfont 50.4 & \fontfamily{bch}\selectfont 43.6 & \fontfamily{bch}\selectfont 35.4 & \fontfamily{bch}\selectfont 39.2 & \fontfamily{bch}\selectfont 20.7 & \fontfamily{bch}\selectfont 60.8 & \fontfamily{bch}\selectfont 41.3 & \fontfamily{bch}\selectfont 43.7 \\

GMNet  & \fontfamily{bch}\selectfont \textbf{92.7} & \fontfamily{bch}\selectfont \textbf{48.0} & \fontfamily{bch}\selectfont \textbf{46.2} & \fontfamily{bch}\selectfont 39.3 & \fontfamily{bch}\selectfont \textbf{69.2} & \fontfamily{bch}\selectfont \textbf{56.0} & \fontfamily{bch}\selectfont \textbf{37.0} & \fontfamily{bch}\selectfont \textbf{45.3} & \fontfamily{bch}\selectfont \textbf{52.6} & \fontfamily{bch}\selectfont \textbf{49.1} & \fontfamily{bch}\selectfont \textbf{50.6} & \fontfamily{bch}\selectfont \textbf{50.6} & \fontfamily{bch}\selectfont 52.0 & \fontfamily{bch}\selectfont \textbf{51.5} & \fontfamily{bch}\selectfont \textbf{24.8} & \fontfamily{bch}\selectfont \textbf{52.6} & \fontfamily{bch}\selectfont \textbf{56.0} & \fontfamily{bch}\selectfont \textbf{40.1} & \fontfamily{bch}\selectfont \textbf{53.9} & \fontfamily{bch}\selectfont \textbf{21.6} & \fontfamily{bch}\selectfont \textbf{70.7} & \fontfamily{bch}\selectfont \textbf{45.8} & \fontfamily{bch}\selectfont \textbf{50.5} \\

\hline
\end{tabular}
\end{table*}

To further verify the robustness and the scalability of the proposed methodology we perform a second set of experiments using an even larger number of parts. 
The results on the Pascal-Part-108 benchmark are reported in Table~\ref{tab:Pascal_part_108}. Even though we can immediately verify a drop in the overall performance, {\revision that is} predictable being the task more complex with respect to the previous scenario with an almost double number of parts, we can appreciate that our framework is able to largely surpass both the baseline and \cite{zhao2019multi}.
It achieves a mIoU of $45.8\%$, outperforming the baseline by $4.5\%$ {\revision and the other compared standard segmentation networks by an even larger margin}. The gain with respect to the main competitor \cite{zhao2019multi} is remarkable with a gap of $2.9\%$ of mIoU.  In this scenario, indeed, most of the previous considerations holds and are even more evident from the results. 
The gain in accuracy is stable across the various classes and parts: the proposed framework significantly wins by large margins on almost every per-object-class mIoU.
 Also for this setup, further results regarding per-part metrics are reported in the supplementary material.

Thanks to the object-level semantic embedding network our model is able to {\revision obtain} accurate segmentation of all the objects with few or no parts inside, such as \textit{boat, bottle, chair, plant} and \textit{sofa}. On these classes, the gain with respect to \cite{zhao2019multi} ranges from $5.4\%$ for the \textit{plant} class to an impressive $15\%$ on the \textit{chair} class.
On the other hand, thanks to the graph matching module, our framework is also able to correctly understand the spatial relationships between small parts, as for example the ones contained in \textit{cat, cow, horse} and \textit{sheep}. Although objects are composed by tiny and difficult parts, the gain with respect to \cite{zhao2019multi} is still significant and ranges between $1.5\%$ on \textit{horse} parts to $11.2\%$ on \textit{cow} ones.

\begin{figure}[tbp]
\setlength\tabcolsep{1.5pt} 
\centering
\begin{tabular}{ccccc}
  RGB & Annotation & Baseline \cite{chen2017rethinking} & BSANet \cite{zhao2019multi} & GMNet (ours) \\

   \includegraphics[width=\sizefiggg\linewidth]{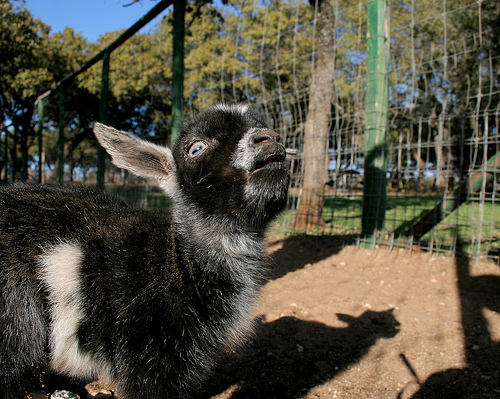} &
  \includegraphics[width=\sizefiggg\linewidth]{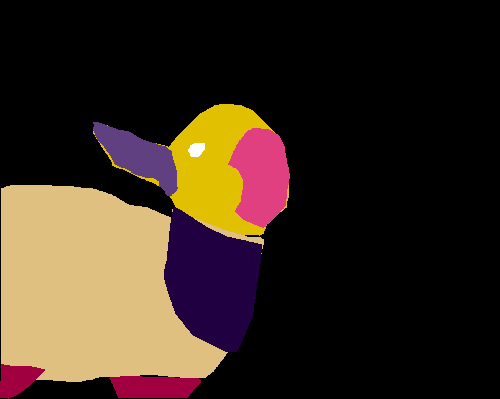} &
  \includegraphics[width=\sizefiggg\linewidth]{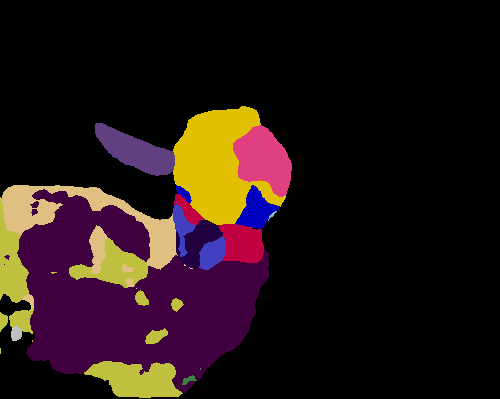} &
  \includegraphics[width=\sizefiggg\linewidth]{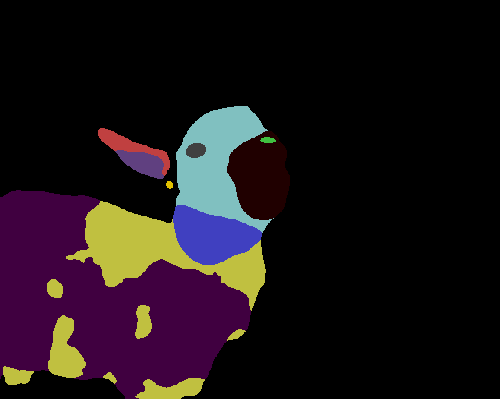} &
  \includegraphics[width=\sizefiggg\linewidth]{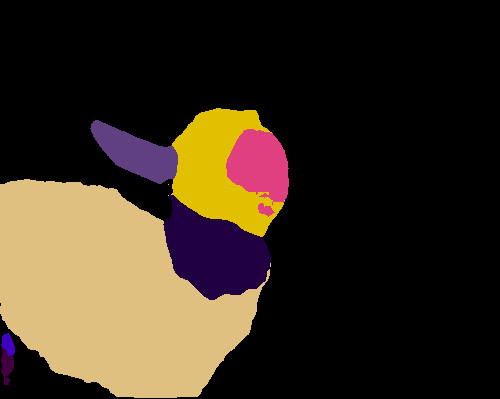} \\

   \includegraphics[width=\sizefiggg\linewidth]{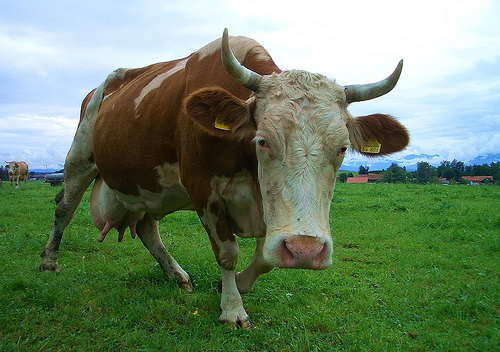} &
  \includegraphics[width=\sizefiggg\linewidth]{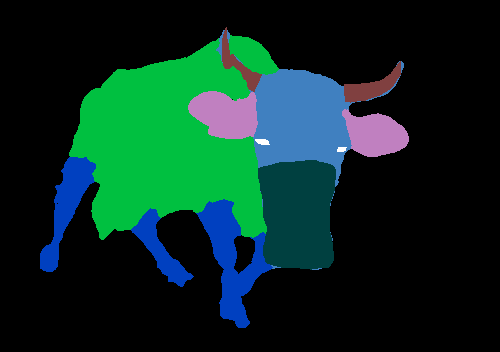} &
  \includegraphics[width=\sizefiggg\linewidth]{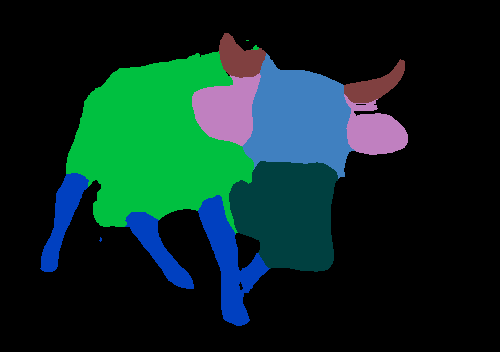} &
  \includegraphics[width=\sizefiggg\linewidth]{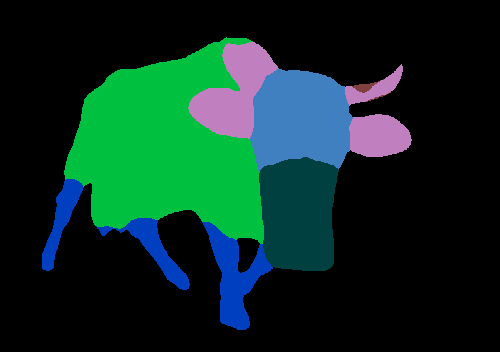} &
  \includegraphics[width=\sizefiggg\linewidth]{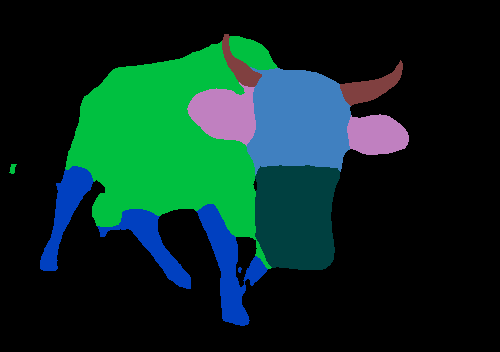} \\
    
   \includegraphics[width=\sizefiggg\linewidth]{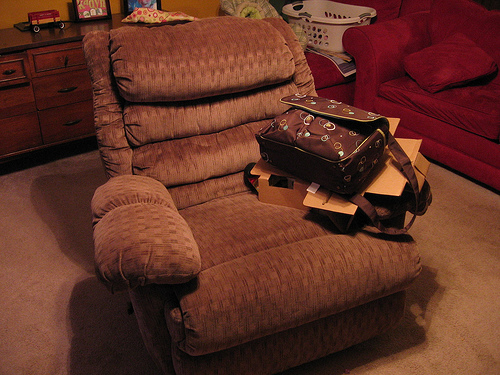} &
  \includegraphics[width=\sizefiggg\linewidth]{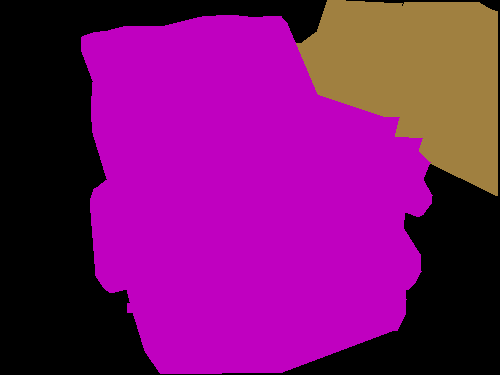} &
  \includegraphics[width=\sizefiggg\linewidth]{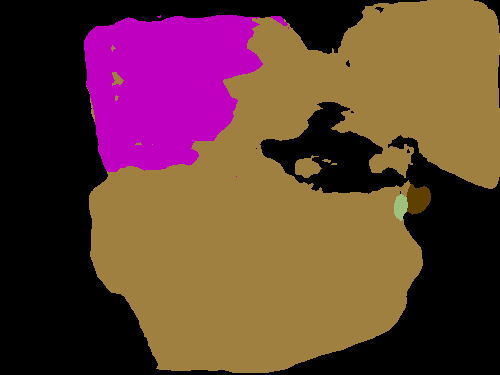} &
  \includegraphics[width=\sizefiggg\linewidth]{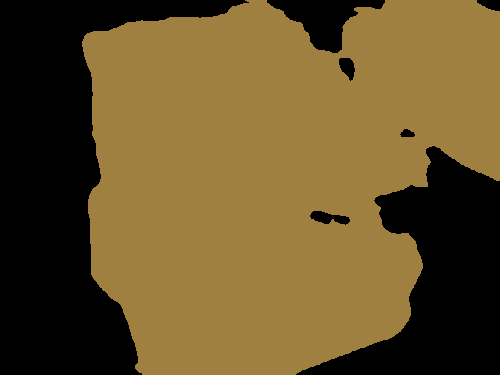} &
  \includegraphics[width=\sizefiggg\linewidth]{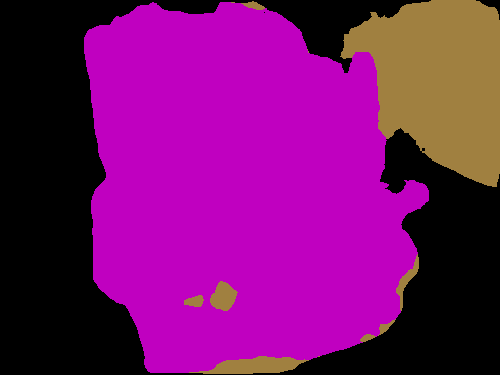} \\
  
  \includegraphics[width=\sizefiggg\linewidth]{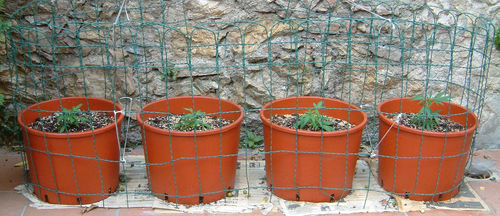} &
  \includegraphics[width=\sizefiggg\linewidth]{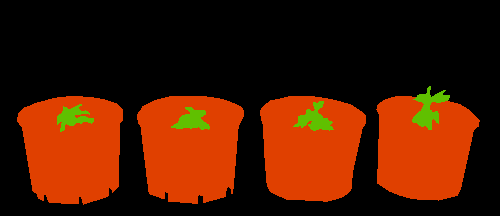} &
  \includegraphics[width=\sizefiggg\linewidth]{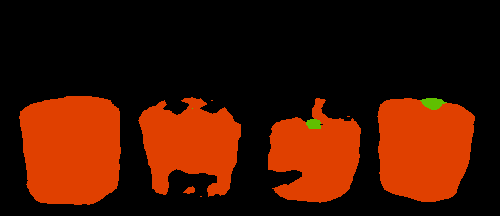} &
  \includegraphics[width=\sizefiggg\linewidth]{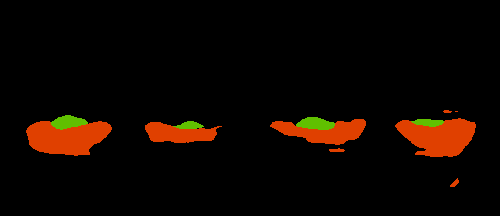} &
  \includegraphics[width=\sizefiggg\linewidth]{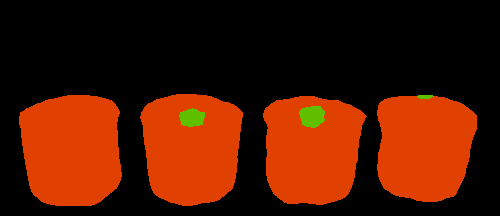} \\
  
 \end{tabular}
\caption{{\revision Segmentation results from} the Pascal-Part-108 dataset (\textit{best viewed in colors}).}
\label{fig:Pascal_part_108}
\end{figure}

The visual results for some sample scenes presented  in Figure~\ref{fig:Pascal_part_108} confirm the numerical evaluation (additional samples are shown in the supplementary material).
 We can appreciate that the proposed method is able to {\revision compute} accurate segmentation maps both when a few elements or many parts coexist in the scene. 
 More in detail, in the first row we can verify the effectiveness of the object-level semantic embedding in conditioning part parsing. The baseline is not able to localize and segment the body and the neck of the sheep. The BSANet approach \cite{zhao2019multi} achieves even worse segmentation and labeling performance. Such methods mislead the sheep with a dog (in the figure  light blue denotes \textit{dog\_head},  light purple  \textit{dog\_neck}, brown \textit{dog\_muzzle} and yellow \textit{dog\_torso}) 
 or with a cat (purple denotes \textit{cat\_torso}). Thanks to the object-level priors, GMNet is able to associate the correct label to each of the parts correctly identifying the sheep as the macro class. 
In the second row, the effect of the graph matching procedure is more evident. The baseline approach tends to overestimate and badly localize the \textit{cow\_horns} (in brown) and BSANet {\revision confuses} the \textit{cow\_horns} with the \textit{cow\_ears} (in pink). GMNet, instead, achieves superior results thanks to the graph module which accounts for proper localization and contour shaping of the various parts. 
In the third row, a scenario with two object-level classes having no sub-parts is reported. Again, we can check how GMNet is able to discriminate between \textit{chair} (in pink) and \textit{sofa} (in light brown).
Finally, in the last row we can appreciate how the two parts of the \textit{potted plant} are correctly segmented by GMNet thanks to the semantic embedding module for what concerns object identification and to the graph matching strategy for what concerns small parts localization.

%% file: sections/ablation.tex
 \subsection{Ablation Studies}
\label{sec:ablation}

In this section we conduct an accurate investigation of the effectiveness of the various modules of the proposed  work on the Pascal-Part-58 dataset.

We start by evaluating the individual impact of the modules and the performance analysis is shown in Table~\ref{tab:ablation_58}.
Let us recall that the baseline architecture (i.e., the Deeplab-v3 network trained directly on the $58$ parts with only the standard cross-entropy loss enabled) achieves a mIoU of $54.4\%$.
The reconstruction loss on the object-level segmentation maps helps in preserving the object-level shapes rearranging parts into object-level classes and allows to improve the  mIoU to $55.2\%$.
The semantic embedding network $\mathcal{S}$ acts as a powerful class-conditioning module to retain object-level semantics when learning parts and allows to obtain a large performance gain: its combination with the reconstruction loss leads to a mIoU of $58.4\%$. 
The addition of the graph matching procedure further boost the final accuracy to $59.0\%$ of mIoU.
To better understand the contribution of this module we also tried a simpler unweighted graph model whose entries are just binary values representing whether two parts are adjacent or not (column $\mathcal{L}_{GM}^u$ in the table). This simplified graph leads to a mIoU of $58.7\%$, lacking some information about the closeness {\revision of} adjacent parts.

Then, we present a more accurate analysis of the impact of the semantic embedding module and the results are summarized in Table~\ref{tab:ablation_S}. 
First of all, the exploitation of the multiple concatenation between features computed by $\mathcal{S}$ and features of $\mathcal{D}_p$ at different resolutions allows object-level embedding at different scales and enhances the scale invariance. Concatenating only the output of $\mathcal{S}$ with the output of $\mathcal{E}_p$ (we refer to this approach with ``single concatenation"), the final mIoU slightly decreases to $58.7\%$.
\begin{table*}[htpb]
\begin{minipage}{0.49\textwidth}
\caption{mIoU ablation results on Pascal-Part-58. $\mathcal{L}_{GM}^u$: graph matching with unweighted graph.}
\label{tab:ablation_58}
\setlength{\tabcolsep}{4pt}
\centering
\begin{tabular}{|ccccc|c|}
\hline
$\mathcal{L}_{CE}$ & $\mathcal{L}_{rec}$  & $\mathcal{S}$ & $\mathcal{L}_{GM}^u$  & $\mathcal{L}_{GM}$ & mIoU \\\hline
\checkmark & & & & & 54.4 \\
\checkmark & \checkmark & & & & 55.2 \\
\checkmark & \checkmark & \checkmark & & & 58.4 \\
\checkmark & \checkmark & \checkmark & \checkmark & & 58.7 \\
\checkmark & \checkmark & \checkmark & & \checkmark & \textbf{59.0} \\
\hline
\end{tabular}
\end{minipage}
\begin{minipage}{0.02\textwidth}
\end{minipage}
\begin{minipage}{0.49\textwidth}
\caption{mIoU on Pascal-Part-58 with different configurations for the object-level semantic embedding. }
\label{tab:ablation_S}
\setlength{\tabcolsep}{6pt}
\centering
\begin{tabular}{|l|c|}
\hline
Method & mIoU \\\hline
Single concatenation & $58.7$ \\
Without $\mathcal{S}$ & $55.7$ \\
$\mathcal{E}_o$ conditioning & $55.7$ \\
GMNet & $\textbf{59.0}$ \\\hline
With objects GT & $65.6$ \\ \hline
\end{tabular}
\end{minipage}
\end{table*}
In order to evaluate the usefulness of exploiting features extracted from a CNN, we compared the proposed framework with a variation directly concatenating the output of $\mathcal{E}_p$ with the object-level predicted segmentation maps $\hat{\mathbf{Y}}_o$ after a proper rescaling (``without $\mathcal{S}$"). This approach leads to a quite low mIoU of $55.7\%$, thus outlining that the embedding network $\mathcal{S}$ is very effective and that a simple stacking of architectures is not the best option for our task.
 Additionally, we considered also the option of directly feeding object-level  features to the part parsing decoder, i.e.,  we tried to concatenate  the output of $\mathcal{E}_o$ with the output of $\mathcal{E}_p$ and feed these features to $\mathcal{D}_p$ (``$\mathcal{E}_o$ conditioning"). 
Conditioning the part parsing with this approach does not bring in sufficient object-level indication and it leads to a mIoU of $55.7\%$, which is significantly lower than the complete proposed framework ($59.0\%$).
 Finally, to estimate an upper limit of the performance gain coming from the semantic embedding module we fed the object-level semantic embedding network $\mathcal{S}$ with object-level ground truth annotations $\mathbf{Y}_o$ (``with objects GT"), instead of the predictions $\hat{\mathbf{Y}}_o$ (notice that the network $\mathcal{A}_o$ has good {\revision performance} but introduces some errors, as it has $71.5\%$ 
of mIoU at object-level). In this case, a mIoU of $65.6\%$ is achieved, showing that there is still room for improvement. 
 
{\revision 
We conclude remarking that GMNet achieves almost always higher accuracy than the starting baseline, even if small and unstructured parts remain the most challenging to be detected. Furthermore, the gain
depends also on the amount of spatial relationships that can be exploited. 
}

%% file: sections/conclusion.tex
\section{Conclusion}
\label{sec:conclusion}
In this paper, we tackled the emerging task of multi-class semantic part segmentation. We propose a novel coarse-to-fine strategy where the features extracted from a semantic segmentation network are enriched with object-level semantics when learning part-level segmentation. Additionally, we designed a novel adjacency graph-based module that aims at matching the relative spatial relationships between ground truth and predicted parts which has shown large improvements particularly on small parts. 
Combining the proposed methodologies we were able to achieve state-of-the-art results in the challenging task of multi-object part parsing both at a moderate scale and  at a larger one. 

Further research will investigate the extension of the proposed modules to other scenarios. We will also consider the explicit embedding into the proposed framework of the edge information coming from part-level and object-level segmentation maps.
Novel graph representations better capturing part relationships and different matching functions will be investigated.

%% file: sections/suppl_58.tex
\section{Additional Results on Pascal-Part-58}
\label{sec:suppl_res58}

We start by analyzing the per-part-class IoU and PA on the Pascal-Part-58 dataset. The results are shown in Table~\ref{tab_suppl:pascal_part_58}, where it is possible to see that the proposed method (GMNet) outperforms the baseline \cite{chen2017rethinking} approach on almost every part both considering the per-part-IoU and the per-part-PA. With respect to BSANet \cite{zhao2019multi}, GMNet can produce clearly higher results on $15$ objects out of $21$ (such as \textit{bottle}, \textit{bus}, \textit{dog}, \textit{sheep},...) and can produce comparable results on $2$ objects (i.e., on \textit{car} ad \textit{cat}).

We can further verify the ranking of the compared methods analyzing the average metrics reported in Table~\ref{tab_suppl:pascal_part_58_mean}. Here, we can appreciate how GMNet is able to outperform both the baseline and BSANet robustly on all the most widely used metrics for semantic segmentation.

\begin{table}[htbp]
\setlength{\tabcolsep}{1.25pt}
\renewcommand{\arraystretch}{1.6}
\scriptsize
  \centering
  \caption{Per-part IoU and PA on the Pascal-Part-58 dataset.}
    \begin{tabular}{|l|cc|cc|cc|c|l|cc|cc|cc|}
    \cline{1-7}\cline{9-15}
         \multicolumn{1}{|c|}{\multirow{2}[0]{*}{Parts Name}}  & \multicolumn{2}{c|}{\textbf{Baseline}} & \multicolumn{2}{c|}{\textbf{BSANet}} & \multicolumn{2}{c|}{\textbf{GMNet}} &  &   \multicolumn{1}{c|}{\multirow{2}[0]{*}{Parts Name}}     & \multicolumn{2}{c|}{\textbf{Baseline}} & \multicolumn{2}{c|}{\textbf{BSANet}} & \multicolumn{2}{c|}{\textbf{GMNet}} \\\cline{2-7} \cline{10-15}
     & \textbf{IoU} & \textbf{PA} & \textbf{IoU} & \textbf{PA} & \textbf{IoU} & \textbf{PA} &   & & \textbf{IoU} & \textbf{PA} & \textbf{IoU} & \textbf{PA} & \textbf{IoU} & \textbf{PA} \\\clineB{1-7}{2} \clineB{9-15}{2}
    background & 91.1 & 96.3 & 91.6 & 96.7 & \textbf{92.7} & \textbf{96.9} & & cow tail & 0.0 & 0.0 & 7.9 & 8.1 & \textbf{8.1} & \textbf{8.4} \\\cdashline{1-7}
    aeroplane body & 66.6 & 79.8 & \textbf{70.0} & \textbf{81.4} & 69.6 & 81.2 & & cow leg & 46.1 & 62.3 & 53.4 & \textbf{67.5} & \textbf{53.5} & 67.2 \\
    aeroplane engine & 25.7 & 31.4 & \textbf{29.1} & \textbf{33.8} & 25.7 & 31.2 & & cow torso & 69.9 & 83.5 & 73.5 & 85.9 & \textbf{77.1} & \textbf{87.8} \\\cdashline{9-15}
    aeroplane wing & 33.5 & 48.2 & \textbf{38.3} & \textbf{49.1} & 34.2 & 46.4 & & dining table & 43.0 & 55.4 & 43.7 & 54.8 & \textbf{51.3} & \textbf{62.6} \\\cdashline{9-15}
    aeroplane stern & 57.1 & 68.2 & \textbf{59.2} & \textbf{72.5} & 57.2 & 70.8 & & dog head & 78.7 & 88.3 & 82.5 & 91.4 & \textbf{85.0} & \textbf{92.7} \\
    aeroplane wheel & 45.4 & 53.3 & \textbf{53.2} & \textbf{62.5} & 46.8 & 53.3 & & dog leg & 48.1 & 59.9 & 53.8 & 63.0 & \textbf{53.8} & \textbf{64.8} \\\cdashline{1-7}
    bike wheel & 78.0 & 88.1 & 78.0 & \textbf{88.6} & \textbf{81.3} & 88.5 & & dog tail & 27.1 & 39.4 & 31.3 & 38.0 & \textbf{31.4} & \textbf{41.5} \\
    bike body & 48.4 & 61.2 & \textbf{53.4} & \textbf{68.4} & 51.5 & 64.2 & & dog torso & 63.7 & 76.8 & 65.7 & 79.7 & \textbf{68.0} & \textbf{81.2} \\\cdashline{1-7}\cdashline{9-15}
    bird head & 64.6 & 72.7 & \textbf{74.0} & \textbf{80.2} & 71.1 & 79.3 & & horse head & 74.7 & 81.7 & \textbf{76.6} & \textbf{83.3} & 73.9 & 80.5 \\
    bird wing & 35.1 & 45.5 & \textbf{39.7} & \textbf{53.2} & 38.6 & 52.9 & & horse tail & 47.0 & 60.4 & \textbf{51.0} & 59.9 & 50.4 & \textbf{62.2} \\
    bird leg & 29.3 & 37.6 & \textbf{34.8} & \textbf{42.6} & 28.7 & 35.4 &  & horse leg & 55.9 & 70.9 & \textbf{61.6} & \textbf{75.8} & 59.3 & 72.9 \\
    bird torso & 66.9 & 83.1 & \textbf{70.9} & \textbf{84.4} & 69.5 & 83.1 & & horse torso & 70.3 & 84.2 & \textbf{74.9} & 86.6 & 73.9 & \textbf{87.4} \\\cdashline{1-7}\cdashline{9-15}
    boat  & 54.4 & 64.8 & 60.2 & 69.6 & \textbf{70.0} & \textbf{78.5} & & mbike wheel & 70.9 & 82.5 & 71.6 & 82.1 & \textbf{73.5} & \textbf{84.0} \\\cdashline{1-7}
    bottle cap & 30.7 & 35.4 & 29.8 & 35.0 & \textbf{33.9} & \textbf{42.5} & & mbike body & 65.1 & 80.9 & 71.5 & 87.7 & \textbf{74.3} & \textbf{87.8} \\\cdashline{9-15}
    bottle body & 68.8 & 78.5 & 68.6 & 74.8 & \textbf{77.6} & \textbf{86.1} & & person head & 83.5 & 91.6 & \textbf{85.0} & \textbf{92.3} & 84.7 & 91.8 \\\cdashline{1-7}
    bus window & 72.7 & 83.7 & 74.8 & 85.9 & \textbf{75.4} & \textbf{86.1} & & person torso & 65.9 & 80.6 & \textbf{68.2} & \textbf{82.7} & 67.0 & 82.3 \\
    bus wheel & 55.3 & 66.3 & 57.1 & 70.1 & \textbf{58.1} & \textbf{72.1} & & person larm & 46.9 & 60.0 & \textbf{52.0} & \textbf{65.6} & 48.6 & 62.8 \\
    bus body & 74.8 & 88.2 & 78.3 & 88.7 & \textbf{79.9} & \textbf{89.8} & & person uarm & 51.5 & 65.8 & \textbf{54.4} & \textbf{68.2} & 52.4 & 66.9 \\\cdashline{1-7}
    car window & 62.6 & 73.9 & \textbf{68.1} & \textbf{78.2} & 64.8 & 77.5 & & person lleg & 38.6 & 51.5 & \textbf{43.5} & \textbf{54.6} & 40.2 & 51.5 \\
    car wheel & 64.8 & 78.1 & 68.5 & 79.7 & \textbf{70.3} & \textbf{79.8} & & person uleg & 43.8 & 60.0 & \textbf{47.4} & \textbf{63.5} & 44.5 & 59.9 \\\cdashline{9-15}
    car light & 46.2 & 54.3 & \textbf{53.7} & \textbf{61.7} & 48.4 & 56.0 & & pplant pot & 45.3 & 61.0 & 53.5 & 64.8 & \textbf{56.0} & \textbf{69.1} \\
    car plate & \textbf{0.0} & \textbf{0.0} & \textbf{0.0} & \textbf{0.0} & \textbf{0.0} & \textbf{0.0} & & pplant plant & 52.4 & 62.1 & \textbf{56.6} & 65.8 & 56.4 & \textbf{66.4} \\\cdashline{9-15}
    car body & 72.1 & 86.4 & 77.0 & \textbf{88.4} & \textbf{77.6} & 88.2 &  & sheep head & 60.9 & 69.3 & 65.4 & 71.3 & \textbf{70.8} & \textbf{79.0} \\\cdashline{1-7}
    cat head & 80.2 & 90.4 & 83.7 & \textbf{92.3} & \textbf{83.8} & 91.6 &  & sheep leg & 8.6 & 11.1 & 11.7 & 16.5 & \textbf{14.3} & \textbf{20.2} \\
    cat leg & 48.6 & \textbf{61.2} & \textbf{50.1} & 58.6 & 49.4 & 59.1 & & sheep torso & 68.3 & 84.4 & 71.6 & 86.1 & \textbf{75.6} & \textbf{88.7} \\\cdashline{9-15}
    cat tail & 40.2 & 51.3 & \textbf{48.8} & 55.6 & 46.0 & \textbf{56.7} & & sofa  & 43.2 & 58.8 & 43.1 & 57.4 & \textbf{56.1} & \textbf{65.0} \\\cdashline{9-15}
    cat torso & 70.3 & 85.7 & 72.6 & \textbf{88.0} & \textbf{73.8} & 87.6 & & train & 79.6 & 86.1 & 82.2 & 90.2 & \textbf{85.0} & \textbf{92.0} \\\cdashline{1-7}\cdashline{9-15}
    chair & 35.4 & 43.3 & 36.5 & 42.7 & \textbf{51.4} & \textbf{63.9} & & tv screen & 69.5 & 76.0 & 73.1 & 78.6 & \textbf{77.0} & \textbf{84.3} \\\cdashline{1-7}
    cow head & 74.3 & 85.6 & 76.4 & 86.0 & \textbf{80.7} & \textbf{87.8} & & tv frame & 45.9 & 56.9 & 49.8 & 60.9 & \textbf{54.1} & \textbf{67.4} \\\cline{1-7}\cline{9-15}
    \end{tabular}%
  \label{tab_suppl:pascal_part_58}%
\end{table}%

\begin{table}
\caption{Comparison in terms of mIoU, mCA and mPA on Pascal-Part-58.}
\label{tab_suppl:pascal_part_58_mean}
\setlength{\tabcolsep}{6pt}
\centering
\begin{tabular}{|l|c|c|c|}
\hline
Method & mIoU & mPA & mCA \\\hline
Baseline \cite{chen2017rethinking} & $54.45$ & $89.86$ & $65.42$ \\
BSANet \cite{zhao2019multi} & $58.15$ & $90.76$ & $68.12$\\
GMNet & $\mathbf{59.04}$ & $\mathbf{91.55}$ & $\mathbf{69.22}$\\\hline
\end{tabular}
\end{table}

\newcommand{\sizefigggg}{0.19}
\begin{figure}[htbp]{}
\setlength\tabcolsep{1.5pt} 
\centering
\begin{tabular}{ccccc}
  RGB & Annotation & Baseline \cite{chen2017rethinking} & BSANet \cite{zhao2019multi} & GMNet (ours) \\
  
      \includegraphics[width=\sizefigggg\linewidth]{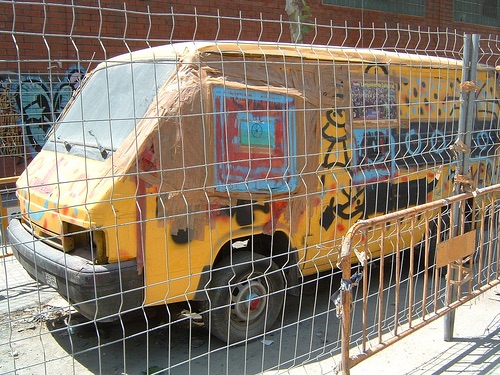} &
  \includegraphics[width=\sizefigggg\linewidth]{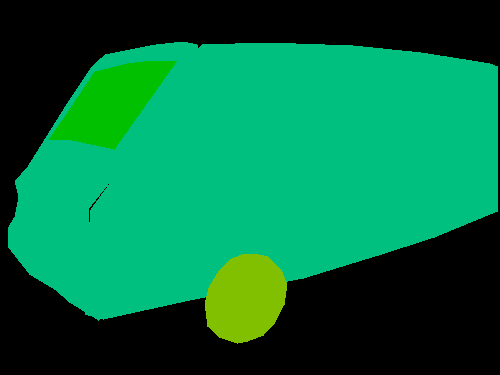} &
  \includegraphics[width=\sizefigggg\linewidth]{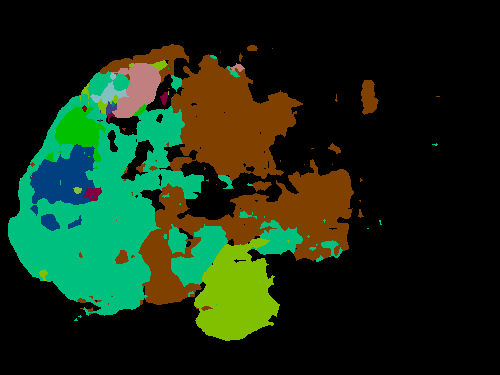} &
  \includegraphics[width=\sizefigggg\linewidth]{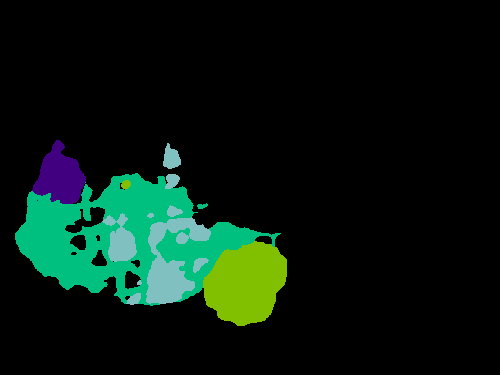} &
  \includegraphics[width=\sizefigggg\linewidth]{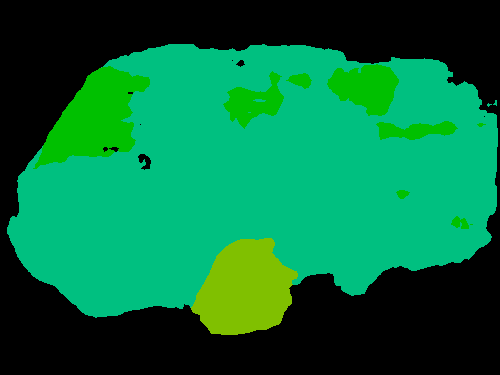} \\
  
    \includegraphics[width=\sizefigggg\linewidth]{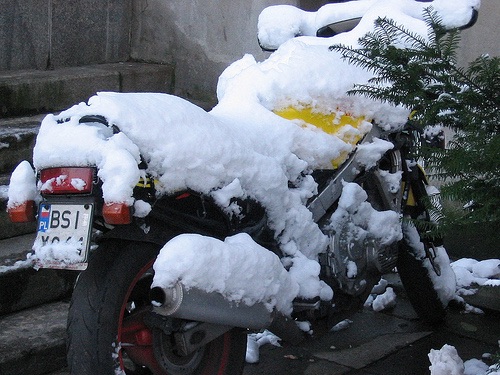} &
  \includegraphics[width=\sizefigggg\linewidth]{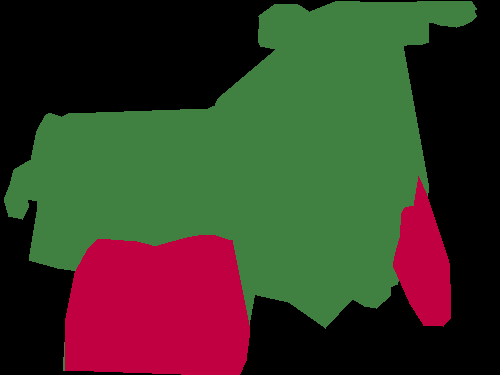} &
  \includegraphics[width=\sizefigggg\linewidth]{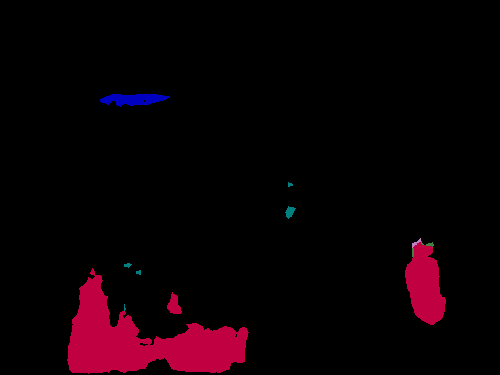} &
  \includegraphics[width=\sizefigggg\linewidth]{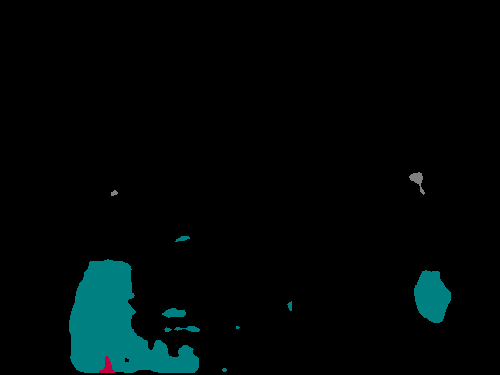} &
  \includegraphics[width=\sizefigggg\linewidth]{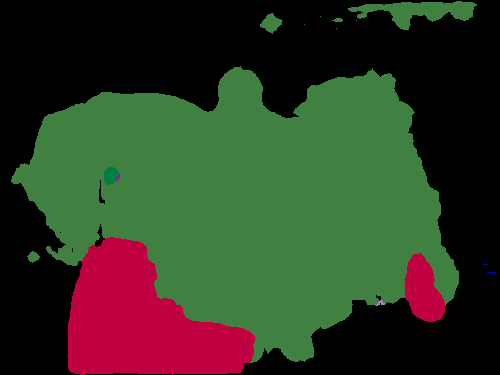} \\
  
    \includegraphics[width=\sizefigggg\linewidth]{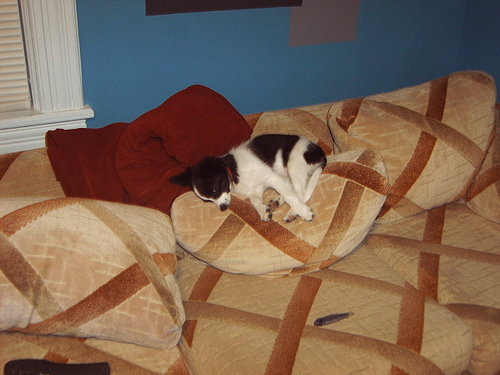} &
  \includegraphics[width=\sizefigggg\linewidth]{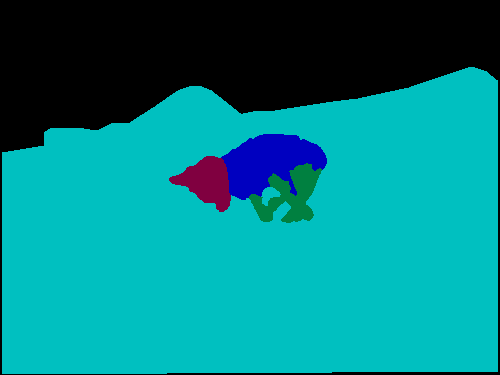} &
  \includegraphics[width=\sizefigggg\linewidth]{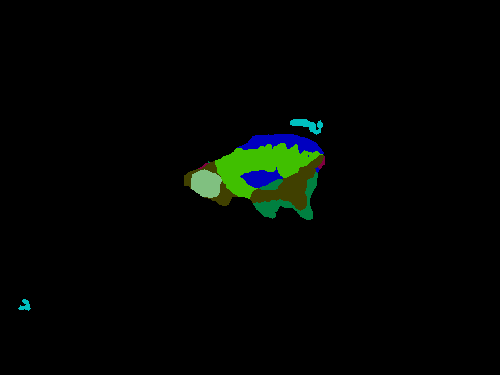} &
  \includegraphics[width=\sizefigggg\linewidth]{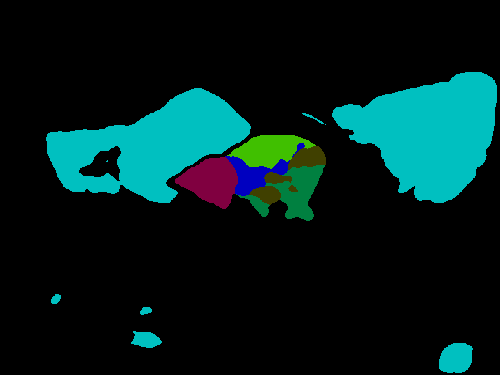} &
  \includegraphics[width=\sizefigggg\linewidth]{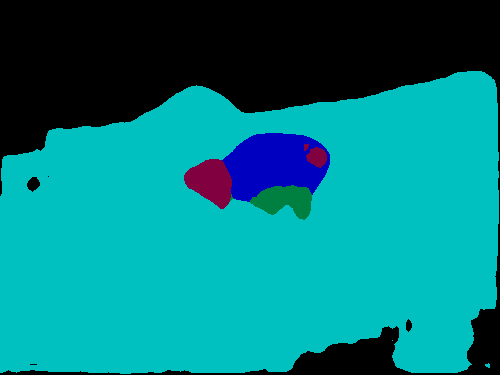} \\
  
    \includegraphics[width=\sizefigggg\linewidth]{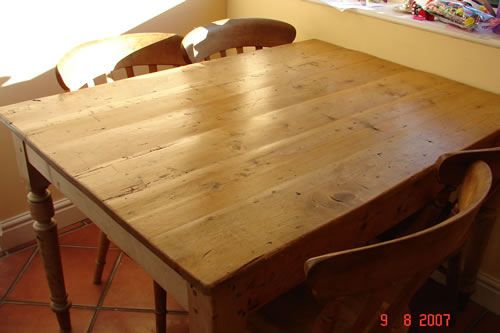} &
  \includegraphics[width=\sizefigggg\linewidth]{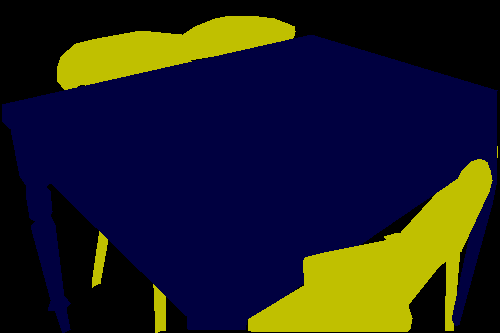} &
  \includegraphics[width=\sizefigggg\linewidth]{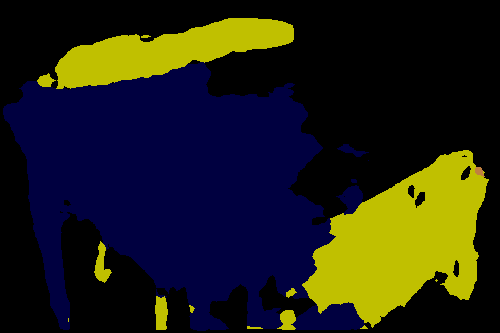} &
  \includegraphics[width=\sizefigggg\linewidth]{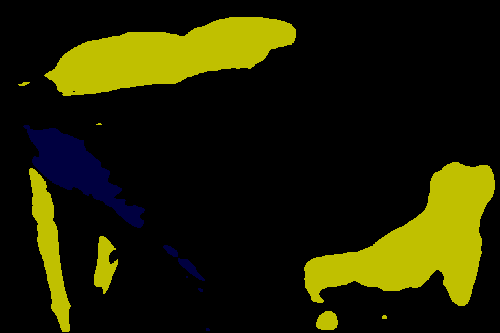} &
  \includegraphics[width=\sizefigggg\linewidth]{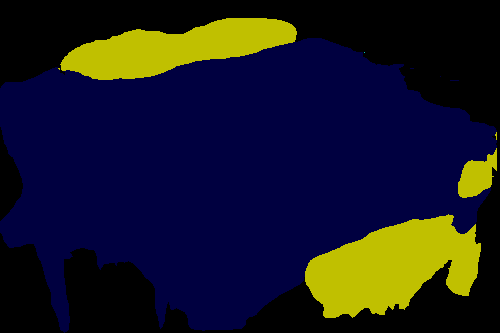} \\
  
      \includegraphics[width=\sizefigggg\linewidth]{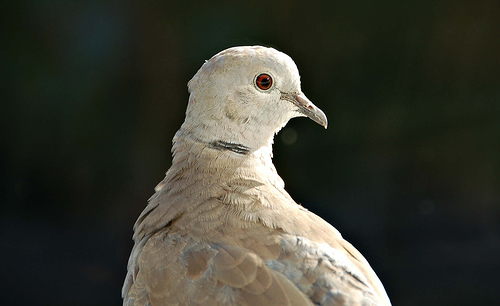} &
  \includegraphics[width=\sizefigggg\linewidth]{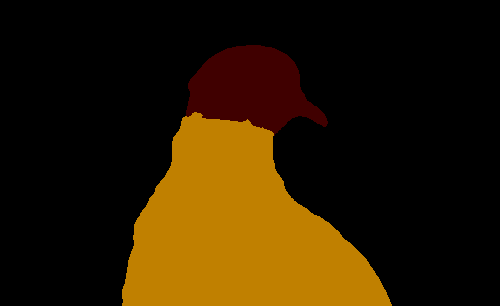} &
  \includegraphics[width=\sizefigggg\linewidth]{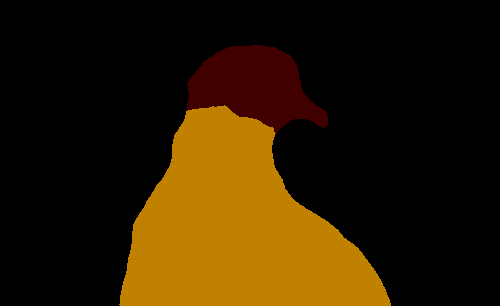} &
  \includegraphics[width=\sizefigggg\linewidth]{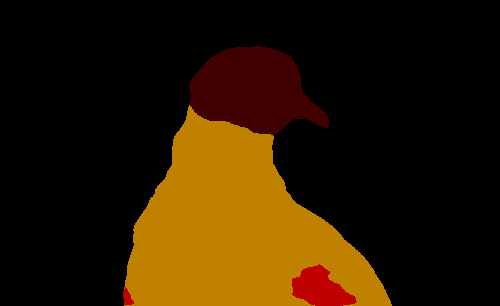} &
  \includegraphics[width=\sizefigggg\linewidth]{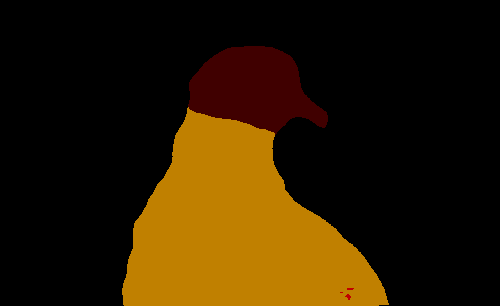} \\

  \includegraphics[width=\sizefigggg\linewidth]{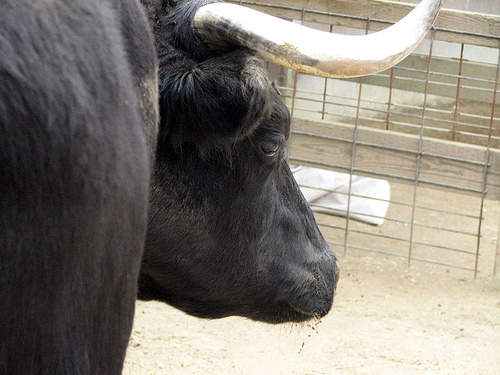} &
  \includegraphics[width=\sizefigggg\linewidth]{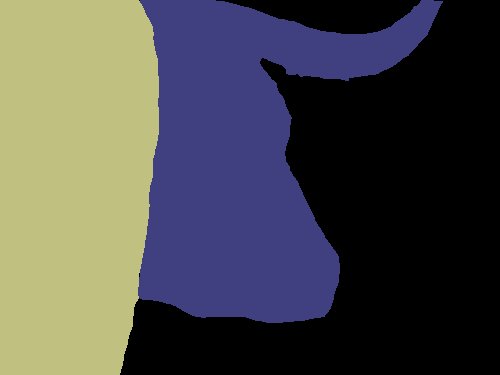} &
  \includegraphics[width=\sizefigggg\linewidth]{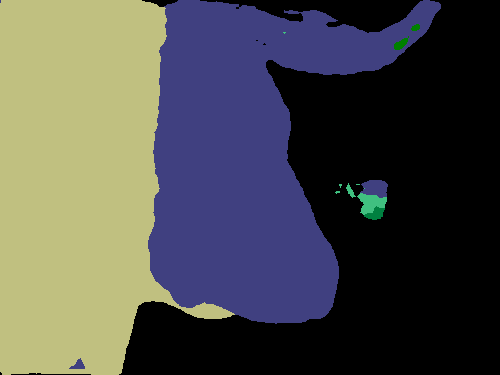} &
  \includegraphics[width=\sizefigggg\linewidth]{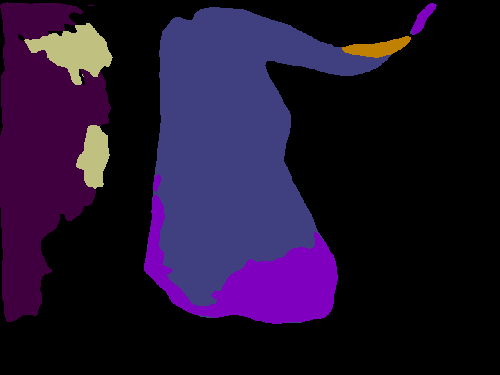} &
  \includegraphics[width=\sizefigggg\linewidth]{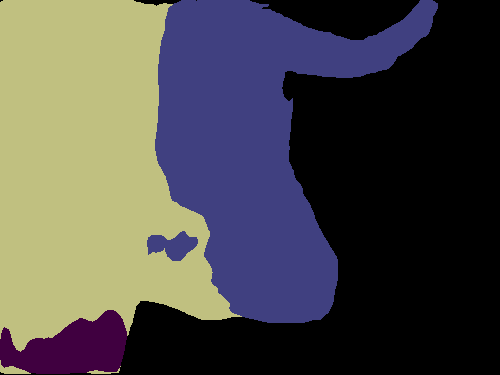} \\
  
  \includegraphics[width=\sizefigggg\linewidth]{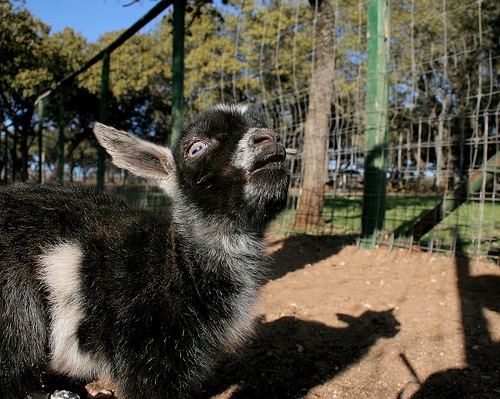} &
  \includegraphics[width=\sizefigggg\linewidth]{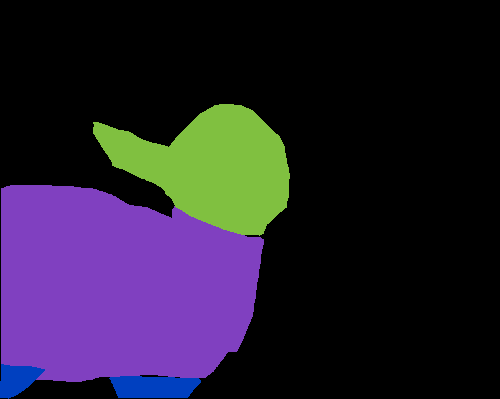} &
  \includegraphics[width=\sizefigggg\linewidth]{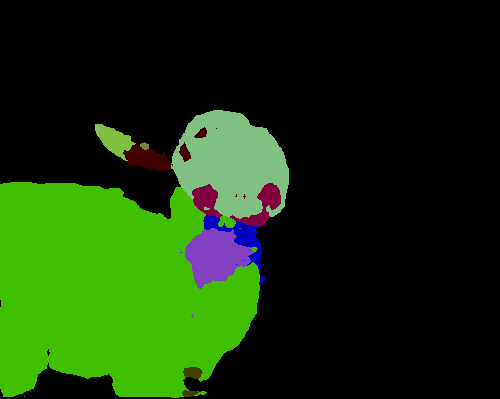} &
  \includegraphics[width=\sizefigggg\linewidth]{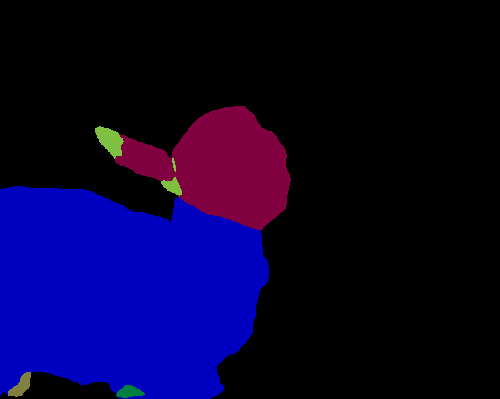} &
  \includegraphics[width=\sizefigggg\linewidth]{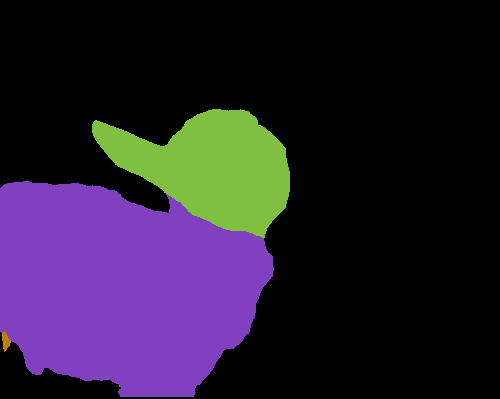} \\

  \includegraphics[width=\sizefigggg\linewidth]{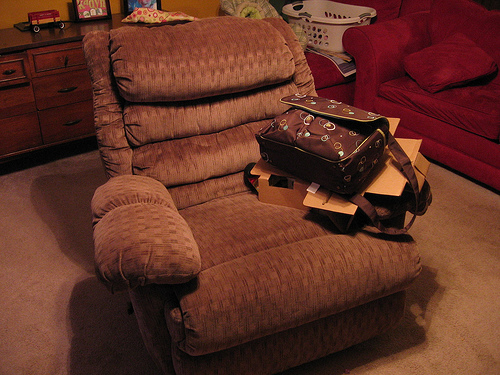} &
  \includegraphics[width=\sizefigggg\linewidth]{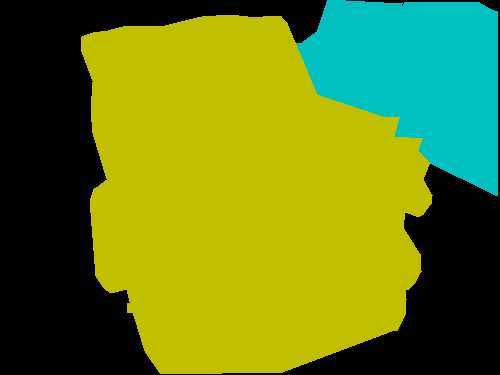} &
  \includegraphics[width=\sizefigggg\linewidth]{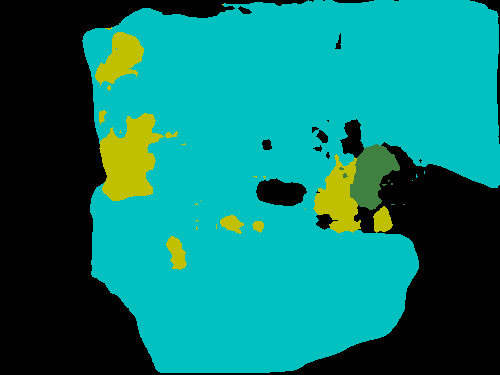} &
  \includegraphics[width=\sizefigggg\linewidth]{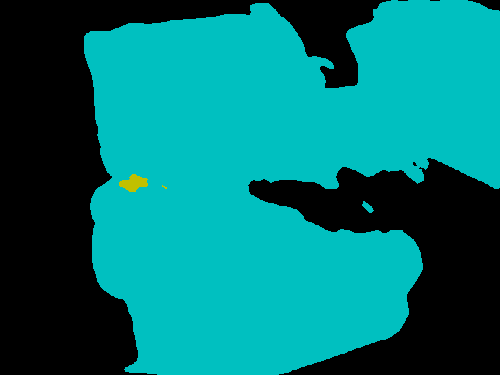} &
  \includegraphics[width=\sizefigggg\linewidth]{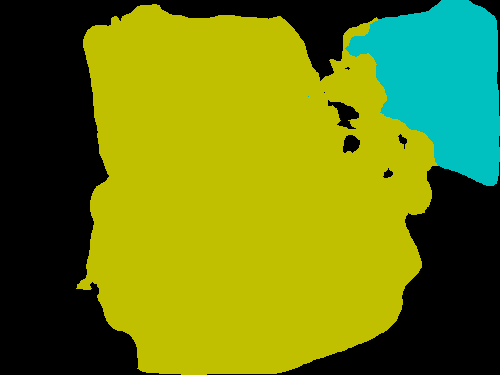} \\

  \includegraphics[width=\sizefigggg\linewidth]{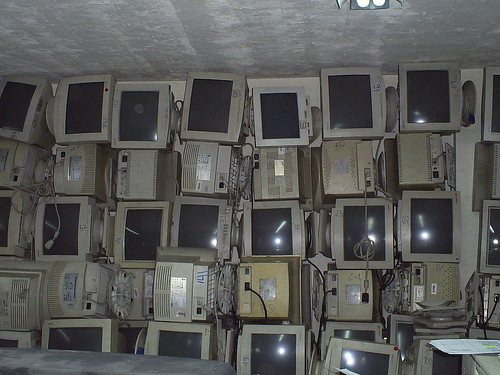} &
  \includegraphics[width=\sizefigggg\linewidth]{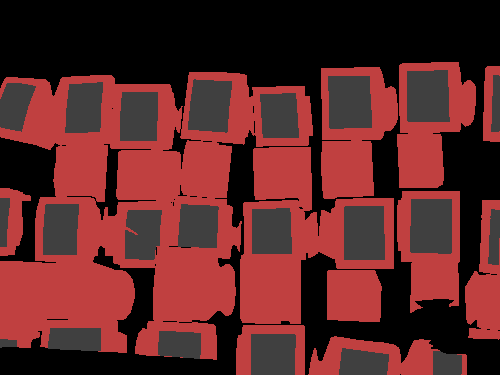} &
  \includegraphics[width=\sizefigggg\linewidth]{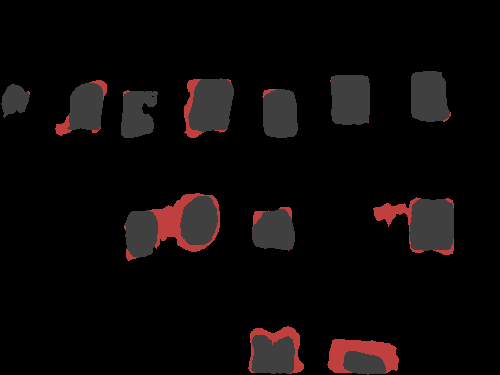} &
  \includegraphics[width=\sizefigggg\linewidth]{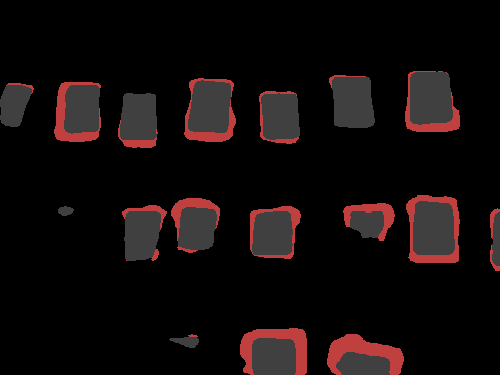} &
  \includegraphics[width=\sizefigggg\linewidth]{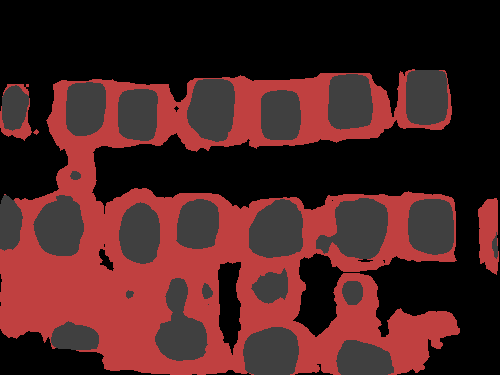} \\
  
       \includegraphics[width=\sizefigggg\linewidth]{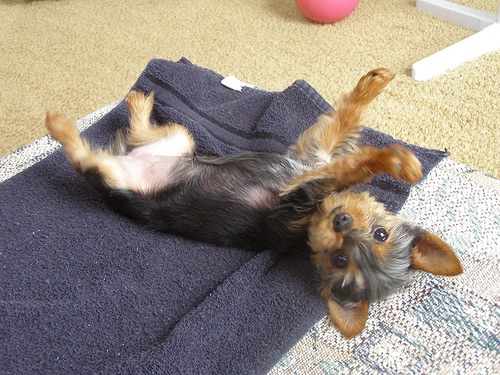} &
  \includegraphics[width=\sizefigggg\linewidth]{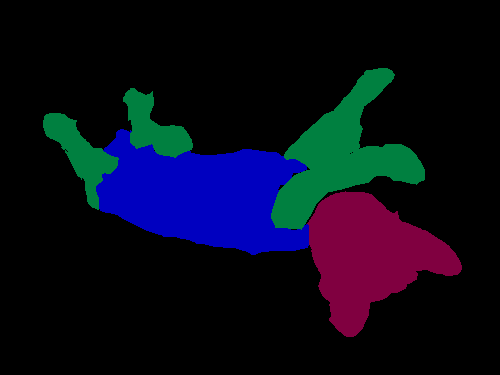} &
  \includegraphics[width=\sizefigggg\linewidth]{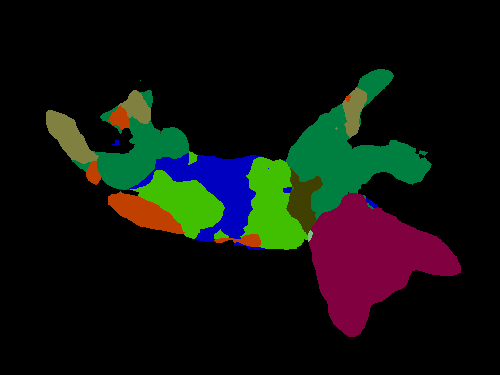} &
  \includegraphics[width=\sizefigggg\linewidth]{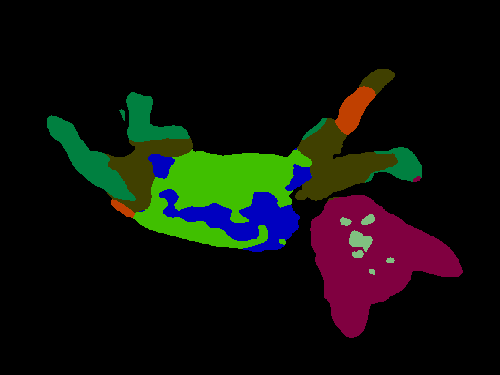} &
  \includegraphics[width=\sizefigggg\linewidth]{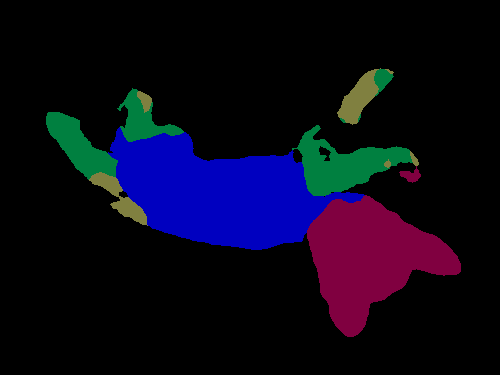} \\

 \end{tabular}
\vspace{-0.1cm}
\caption{Qualitative results on some sample scenes on the Pascal-Part-58 dataset (\textit{best viewed in colors}).}
\label{fig_suppl:Pascal_part_58}
\end{figure}

Then, we proceed to analyze some additional qualitative results as reported in Figure~\ref{fig_suppl:Pascal_part_58}. The effects of the two main components of our work, namely the object-level semantic embedding network $\mathcal{S}$ and the graph matching module, are clearly visible in the images.
The effect of the semantic embedding network is evident in the last $5$ rows, where object-level conditioning helps the part-level decoder to accurately segment and label the parts. For instance, in the last row both the baseline and BSANet mislead the dog's parts with cat's parts, while GMNet is able to avoid this error. In row $6$, BSANet confuses cow's parts with sheep's parts. In row $7$, the baseline confuses sheep's parts with cat's ones and BSANet with dog's parts. GMNet is able to correctly deal with these situations thanks to the object-level guidance.

The effect of the graph matching module is more appreciable on small parts. For example, we can verify its efficacy in the third row and in the last row. In row $3$, both the baseline and BSANet mislead the dog's parts with cat's ones and also their localization is highly imprecise. From one hand, the semantic embedding network corrects the first issue, while the second (i.e., bad localization) is addressed by graph matching. In the last row, graph matching between different reciprocal spatial relationship among parts helps to correctly place the dog's parts.

Moreover, in very challenging images (such as rows $1$ to $4$), where both the baseline and BSANet partially or completely miss some classes, our method generate superior quality segmentation maps. For instance, in row $1$ a vehicle behind a metal grid is being correctly identified and quite well localized in all its parts thanks to the semantic embedding module and to graph matching. The combination of the two modules is also helpful in row $2$, where a motorbike covered with snow is being well recognized by our framework. In row $3$ we identify the sofa and in row $4$ the table, with higher accuracy than the compared methods.

%% file: sections/suppl_108.tex
\vspace{1.5cm}

\section{Additional Results on Pascal-Part-108}
\label{sec:suppl_res108}

In this section we present some additional results for the Pascal-Part-108 dataset. The per-part-IoU and per-part-PA are reported in Table~\ref{tab_suppl:pascal_part_108}, where we can notice that the gap between the proposed framework and the compared methods is significantly larger than for the  Pascal-Part-58 dataset. 
GMNet achieves higher accuracy than the competitors on almost all the parts. In particular, our framework is able to outperform BSANet \cite{zhao2019multi} in $19$ out of $21$ object-level classes both with many parts within them (such as \textit{aeroplane}, \textit{bus}, \textit{cat}, \textit{dog}, \textit{person}, \textit{sheep},...) and with no or few parts within them (such as \textit{boat}, \textit{bottle}, \textit{chair}, \textit{sofa}, \textit{tv},..).

The mean accuracy results are shown in Table~\ref{tab_suppl:pascal_part_108_mean} where we can verify that our method clearly outperforms both the baseline \cite{chen2017rethinking} and BSANet \cite{zhao2019multi} on all the most popular metrics used to evaluate semantic segmentation architectures. Hence, we prove the robustness of our framework to different evaluation criteria and to different datasets. Additionally, we argue that the proposed framework is able to scale well to even larger sets of parts.

\begin{table}[htbp]
\scriptsize
\setlength{\tabcolsep}{1.25pt}
\renewcommand{\arraystretch}{1.15}
  \centering
  \caption{Per-part IoU and PA on the Pascal-Part-108 dataset.}
    \begin{tabular}{|l|cc|cc|cc|c|l|cc|cc|cc|}
    \cline{1-7}\cline{9-15}
     \multicolumn{1}{|c|}{\multirow{2}[0]{*}{Parts Name}}  & \multicolumn{2}{c|}{\textbf{Baseline}} & \multicolumn{2}{c|}{\textbf{BSANet}} & \multicolumn{2}{c|}{\textbf{GMNet}} & & \multicolumn{1}{c|}{\multirow{2}[0]{*}{Parts Name}}  & \multicolumn{2}{c|}{\textbf{Baseline}} & \multicolumn{2}{c|}{\textbf{BSANet}} & \multicolumn{2}{c|}{\textbf{GMNet}} \\\cline{2-7} \cline{10-15}
          & \textbf{IoU} & \textbf{PA} & \textbf{IoU} & \textbf{PA} & \textbf{IoU} & \textbf{PA} &  &      & \textbf{IoU} & \textbf{PA} & \textbf{IoU} & \textbf{PA} & \textbf{IoU} & \textbf{PA} \\\clineB{1-7}{2} \clineB{9-15}{2}
    background & 90.9 & \textbf{97.2} & 91.6 & 97.1 & \textbf{92.7} & 97.0 &  & dining table & 33.0 & 40.2 & 45.9 & 59.7 & \textbf{50.6} & \textbf{62.3} \\\cdashline{1-7}\cdashline{9-15}
    aero body & 61.9 & 72.3 & \textbf{68.2} & 77.6 & 61.9 & \textbf{82.6} & &  dog head & 60.5 & 75.5 & 63.8 & 78.2 & \textbf{64.0} & \textbf{78.9} \\
    aero stern & 53.2 & 68.4 & 54.2 & 65.3 & \textbf{57.4} & \textbf{71.0} & &  dog reye & 50.1 & 61.4 & 54.1 & 61.4 & \textbf{54.7} & \textbf{64.7} \\
    aero rwing & 28.9 & 39.8 & 33.1 & \textbf{46.5} & \textbf{34.3} & 46.0 & &  dog rear & 54.0 & 69.4 & \textbf{57.2} & 73.4 & 56.8 & \textbf{73.9} \\
    aero engine & 24.7 & 29.0 & 26.5 & 32.0 & \textbf{27.2} & \textbf{32.6} & &  dog nose & 63.5 & 75.0 & \textbf{66.3} & 74.3 & 66.0 & \textbf{76.8} \\
    aero wheel & 40.9 & 46.8 & 44.5 & 49.6 & \textbf{51.5} & \textbf{61.3} &  & dog torso & 58.4 & 74.6 & 62.3 & 78.4 & \textbf{63.2} & \textbf{79.1} \\\cdashline{1-7}
    bike fwheel & 78.4 & 85.7 & 75.3 & 86.7 & \textbf{80.2} & \textbf{87.8} &  & dog neck & 27.1 & 35.4 & 26.2 & 30.8 & \textbf{28.1} & \textbf{35.5} \\
    bike saddle & 34.1 & 39.8 & 31.0 & 31.9 & \textbf{38.0} & \textbf{43.2} & &  dog rfleg & 39.2 & 50.6 & 42.4 & 53.5 & \textbf{43.7} & \textbf{55.8} \\
    bike handlebar & \textbf{23.3} & \textbf{26.1} & 20.6 & 22.8 & 22.4 & 25.9 & &  dog rfpaw & 39.4 & 47.9 & \textbf{44.2} & 51.7 & 43.7 & \textbf{52.9} \\
    bike chainwheel & 42.3 & 50.4 & 36.5 & 41.6 & \textbf{44.1} & \textbf{57.0} &  & dog tail & 24.7 & 37.8 & \textbf{34.9} & \textbf{42.3} & 30.8 & 41.4 \\\cdashline{1-7}
    birds head & 51.5 & 61.3 & \textbf{66.4} & \textbf{78.0} & 65.3 & 77.7 & &  dog muzzle & 65.1 & 76.1 & \textbf{69.4} & \textbf{82.3} & 68.9 & 80.4 \\\cdashline{9-15}
    birds beak & 40.4 & 49.5 & \textbf{47.1} & \textbf{54.6} & 44.3 & 54.0 & &  horse head & 54.4 & 67.0 & \textbf{57.1} & \textbf{68.9} & 55.9 & 68.3 \\
    birds torso & 61.7 & 77.9 & \textbf{65.2} & 79.4 & 64.8 & \textbf{82.6} & &  horse rear & 49.7 & 58.1 & 51.1 & 56.5 & \textbf{52.2} & \textbf{65.6} \\
    birds neck & 27.5 & 32.2 & \textbf{39.1} & \textbf{50.1} & 28.4 & 35.7 & &  horse muzzle & 61.3 & 68.7 & \textbf{65.2} & \textbf{74.0} & 62.9 & 69.5 \\
    birds rwing & 35.9 & 50.4 & \textbf{39.3} & \textbf{53.7} & 37.2 & 50.1 & &  horse torso & 56.7 & 75.9 & 59.5 & 75.9 & \textbf{60.7} & \textbf{84.3} \\
    birds rleg & 23.5 & 28.6 & \textbf{26.5} & 32.2 & 23.8 & \textbf{32.8} & &  horse neck & 42.1 & 51.3 & \textbf{49.6} & \textbf{64.8} & 47.2 & 55.8 \\
    birds rfoot & 13.9 & 16.3 & 11.6 & 12.7 & \textbf{17.7} & \textbf{22.5} & &  horse rfuleg & 54.1 & 68.5 & \textbf{57.0} & \textbf{71.8} & 56.4 & 70.9 \\
    birds tail & 28.1 & 39.2 & \textbf{33.0} & 44.1 & 32.5 & \textbf{46.1} &  & horse tail & 48.1 & 63.5 & 47.6 & 54.5 & \textbf{51.4} & \textbf{64.4} \\\cdashline{1-7}
    boat  & 53.7 & 60.3 & 61.4 & 71.5 & \textbf{69.2} & \textbf{77.8} &  & horse rfho & 24.1 & 31.4 & 12.9 & 13.7 & \textbf{25.3} & \textbf{32.7} \\\cdashline{1-7}\cdashline{9-15}
    bottle cap & 30.4 & 35.0 & 26.2 & 30.0 & \textbf{33.4} & \textbf{40.0} & &  mbike fwheel & 69.6 & 78.9 & 69.3 & 80.4 & \textbf{73.6} & \textbf{83.3} \\
    bottle body & 63.7 & 69.5 & 71.5 & 78.3 & \textbf{78.7} & \textbf{88.3} & &  mbike hbar & \textbf{0.0} & \textbf{0.0} & 0.0 & 0.0 & 0.0 & 0.0 \\\cdashline{1-7}
    bus rightside & 70.8 & 85.3 & 73.0 & 83.7 & \textbf{75.7} & \textbf{88.4} &  & mbike saddle & 0.0 & 0.0 & 0.0 & 0.0 & \textbf{0.8} & \textbf{0.8} \\
    bus roofside & 7.5 & 7.7 & 0.3 & 0.3 & \textbf{13.5} & \textbf{14.4} &  & mbike hlight & 25.8 & \textbf{32.8} & 10.6 & 11.2 & \textbf{28.5} & 32.4 \\\cdashline{9-15}
    bus mirror & 2.1 & 2.2 & 0.3 & 0.3 & \textbf{6.6} & \textbf{7.6} &  & person head & 68.2 & 81.9 & \textbf{69.7} & 82.2 & 69.3 & \textbf{82.7} \\
    bus fliplate & \textbf{0.0} & \textbf{0.0} & \textbf{0.0} & \textbf{0.0} & \textbf{0.0} & \textbf{0.0} &  & person reye & 35.1 & 39.3 & \textbf{41.3} & \textbf{46.3} & 38.7 & 43.9 \\
    bus door & \textbf{40.1} & 51.2 & 37.2 & \textbf{53.2} & 38.1 & 47.3 & &  person rear & 37.4 & 46.0 & \textbf{41.9} & 49.4 & 41.4 & \textbf{51.5} \\
    bus wheel & \textbf{54.8} & 65.5 & 53.1 & 63.9 & 56.7 & \textbf{69.4} &  & person nose & 53.0 & 62.1 & 54.3 & 63.1 & \textbf{56.7} & \textbf{67.5} \\
    bus headlight & \textbf{25.6} & 28.3 & 19.9 & 20.8 & 30.4 & \textbf{34.2} & &  person mouth & 48.9 & 56.9 & 49.5 & 54.9 & \textbf{51.3} & \textbf{60.8} \\
    bus window & 71.8 & 85.2 & 73.5 & 86.4 & \textbf{74.6} & \textbf{87.4} &  & person hair & 70.8 & 83.3 & \textbf{72.3} & \textbf{85.9} & 71.8 & 83.9 \\\cdashline{1-7}
    car rightside & 64.0 & 78.0 & 67.9 & 81.2 & \textbf{70.5} & \textbf{84.5} & &  person torso & 63.4 & 79.1 & 64.3 & 78.3 & \textbf{65.2} & \textbf{80.9} \\
    car roofside & 21.0 & 25.4 & 16.1 & 17.6 & \textbf{22.3} & \textbf{26.6} &  & person neck & 49.7 & 63.8 & 50.9 & 65.1 & \textbf{51.2} & \textbf{65.3} \\
    car fliplate & \textbf{0.0} & \textbf{0.0} & \textbf{0.0} & \textbf{0.0} & \textbf{0.0} & \textbf{0.0} &  & person ruarm & 54.7 & 68.6 & 55.7 & 70.2 & \textbf{57.4} & \textbf{71.3} \\
    car door & 41.4 & 52.5 & 39.6 & 49.0 & \textbf{42.3} & \textbf{53.5} & &  person rhand & 43.0 & 55.4 & \textbf{47.4} & \textbf{57.6} & 44.1 & 56.8 \\
    car wheel & 65.8 & 74.5 & 64.0 & 76.6 & \textbf{70.2} & \textbf{80.0} & &  person ruleg & 50.8 & 66.0 & 52.3 & 67.1 & \textbf{53.0} & \textbf{67.9} \\
    car headlight & 42.9 & 48.4 & \textbf{49.4} & \textbf{59.7} & 46.4 & 54.4 & &  person rfoot & 29.8 & 38.9 & 28.9 & 32.4 & \textbf{31.3} & \textbf{39.8} \\\cdashline{9-15}
    car window & 61.0 & 75.5 & \textbf{66.5} & \textbf{82.4} & 65.0 & 79.0 & &  pplant pot & 43.6 & 54.5 & 50.6 & 58.9 & \textbf{56.0} & \textbf{69.0} \\\cdashline{1-7}
    cat head & 73.9 & 87.3 & 75.6 & 88.5 & \textbf{77.5} & \textbf{88.5} & &  pplant plant & 42.9 & 48.8 & 55.5 & \textbf{68.7} & \textbf{56.6} & 66.6 \\\cdashline{9-15}
    cat reye & 58.8 & 69.0 & 62.0 & 71.1 & \textbf{62.8} & \textbf{71.8} & &  sheep head & 45.6 & 56.9 & 47.0 & 58.0 & \textbf{54.0} & \textbf{66.9} \\
    cat rear & 65.5 & 77.7 & 66.8 & 77.1 & \textbf{67.1} & \textbf{78.8} & &  sheep rear & 43.2 & 53.0 & \textbf{47.7} & 56.6 & 45.3 & \textbf{58.2} \\
    cat nose & 40.3 & 49.1 & 41.2 & 45.8 & \textbf{46.3} & \textbf{56.2} & &  sheep muzzle & 58.2 & 67.0 & 61.1 & 72.4 & \textbf{64.9} & \textbf{74.7} \\
    cat torso & 64.2 & 81.4 & 66.8 & 84.2 & \textbf{68.7} & \textbf{86.0} & &  sheep rhorn & 3.0 & 3.6 & 0.0 & 0.0 & \textbf{5.4} & \textbf{6.6} \\
    cat neck & 22.8 & 33.8 & 19.8 & 25.0 & 24.4 & \textbf{34.1} &  & sheep torso & 62.6 & 78.0 & 66.4 & 83.6 & \textbf{68.8} & \textbf{86.3} \\
    cat rfleg & 36.5 & 48.5 & 38.5 & 49.2 & \textbf{39.1} & \textbf{50.0} &  & sheep neck & 26.9 & 38.1 & 25.3 & \textbf{41.2} & \textbf{30.3} & 41.0 \\
    cat rfpaw & 40.6 & 50.2 & \textbf{43.4} & \textbf{51.5} & 41.7 & 50.7 &  & sheep rfuleg & 8.6 & 10.6 & \textbf{17.4} & \textbf{24.5} & 11.7 & 14.7 \\
    cat tail & 40.2 & 52.2 & 42.6 & 49.5 & \textbf{45.8} & \textbf{57.0} & &  sheep tail & 6.7 & 7.4 & 1.1 & 1.1 & \textbf{9.1} & \textbf{11.5} \\\cdashline{1-7}\cdashline{9-15}
    chair & 35.4 & 42.3 & 34.1 & 38.4 & \textbf{49.1} & \textbf{60.4} &  & sofa  & 39.2 & 50.7 & 44.5 & 56.9 & \textbf{53.9} & \textbf{66.1} \\\cdashline{1-7}\cdashline{9-15}
    cow head & 51.2 & 65.5 & 58.2 & 74.2 & \textbf{63.8} & \textbf{74.9} &  & train head & 5.3 & \textbf{6.4} & \textbf{5.6} & 6.4 & 4.5 & 5.3 \\
    cow rear & 51.2 & 68.5 & 53.0 & 72.9 & \textbf{60.0} & \textbf{75.1} &  & train hrightside & 61.9 & 77.3 & \textbf{63.5} & \textbf{84.0} & 60.8 & 83.1 \\
    cow muzzle & 61.2 & 77.6 & 67.2 & 81.9 & \textbf{74.9} & \textbf{86.7} &  & train hroofside & \textbf{23.0} & \textbf{28.0} & 13.7 & 17.0 & 21.1 & 26.3 \\
    cow rhorn & 28.8 & 35.0 & 10.1 & 10.2 & \textbf{44.0} & \textbf{50.6} &  & train headlight & \textbf{0.0} & \textbf{0.0} & \textbf{0.0} & \textbf{0.0} & \textbf{0.0} & \textbf{0.0} \\
    cow torso & 63.4 & 78.6 & 69.9 & 85.8 & \textbf{73.2} & \textbf{87.2} & &  train coach & 28.6 & 33.6 & \textbf{42.0} & \textbf{47.8} & 31.4 & 37.9 \\
    cow neck & 9.5 & 12.7 & 7.3 & 7.9 & \textbf{20.3} & \textbf{25.9} & &  train crightside & 15.6 & 24.5 & \textbf{19.0} & 30.6 & 14.9 & \textbf{33.8} \\
    cow rfuleg & 46.5 & 60.0 & 49.7 & 61.4 & \textbf{54.8} & \textbf{70.7} & &  train croofside & 10.8 & 11.9 & 1.0 & 1.0 & \textbf{18.1} & \textbf{22.6} \\\cdashline{9-15}
    cow tail & 6.5 & 7.3 & 0.1 & 0.1 & \textbf{13.6} & \textbf{14.9} &  & tv screen & 60.8 & 71.3 & 66.3 & 79.5 & \textbf{70.7} & \textbf{82.9} \\\cline{1-7}\cline{9-15}
    \end{tabular}%
  \label{tab_suppl:pascal_part_108}%
\end{table}%

\begin{figure}[htbp]{}
\setlength\tabcolsep{1.5pt} 
\centering
\begin{tabular}{ccccc}
  RGB & Annotation & Baseline \cite{chen2017rethinking} & BSANet \cite{zhao2019multi} & GMNet (ours) \\
  
    \includegraphics[width=\sizefiggg\linewidth]{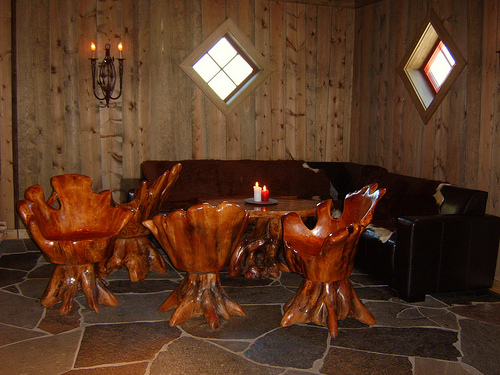} &
  \includegraphics[width=\sizefigggg\linewidth]{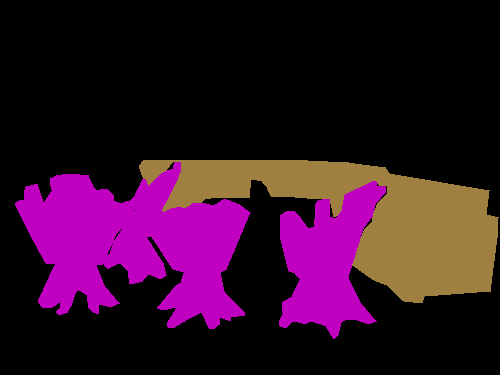} &
  \includegraphics[width=\sizefigggg\linewidth]{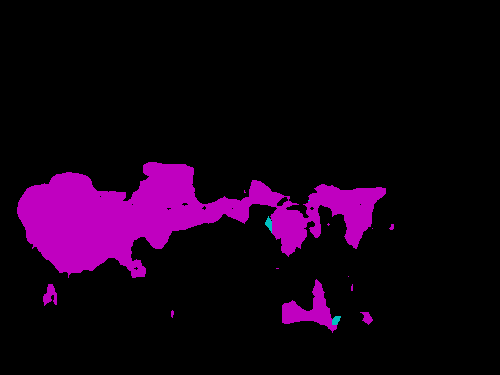} &
  \includegraphics[width=\sizefigggg\linewidth]{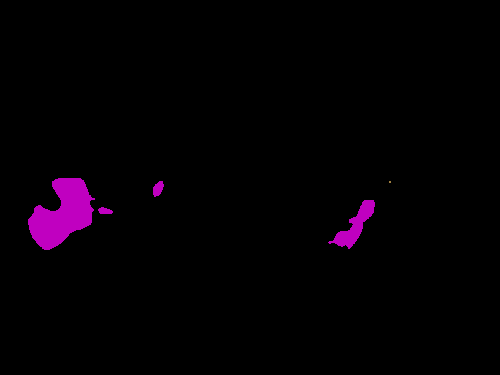} &
  \includegraphics[width=\sizefigggg\linewidth]{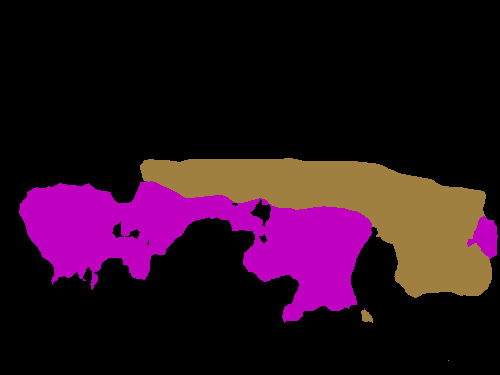} \\
  
     \includegraphics[width=\sizefigggg\linewidth]{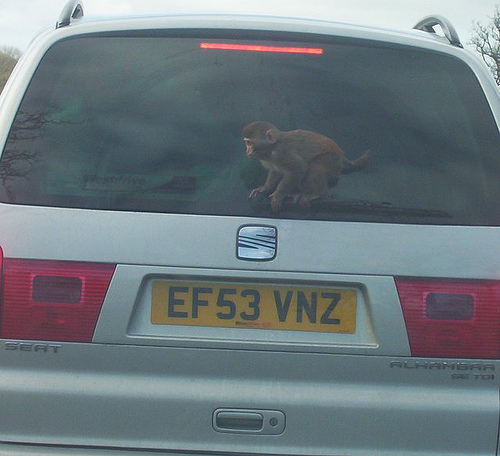} &
  \includegraphics[width=\sizefigggg\linewidth]{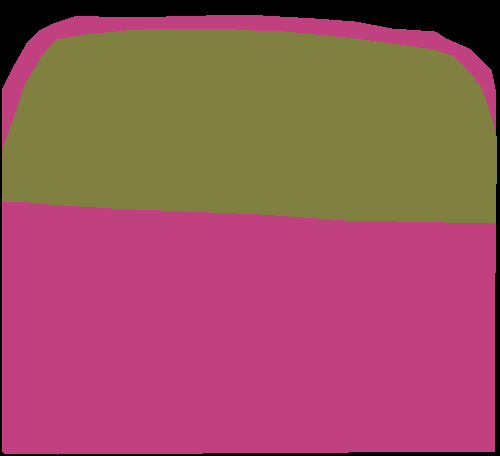} &
  \includegraphics[width=\sizefigggg\linewidth]{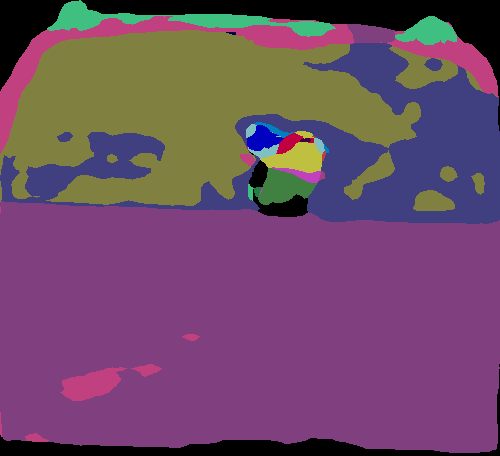} &
  \includegraphics[width=\sizefigggg\linewidth]{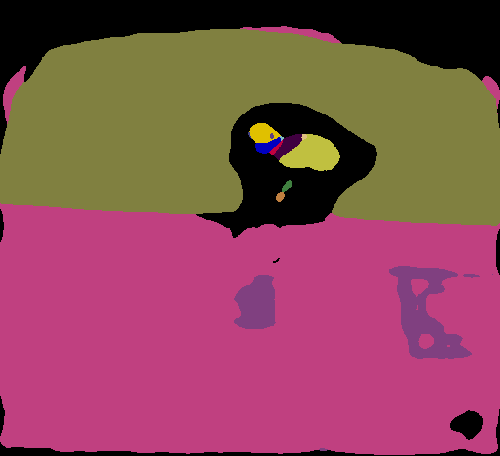} &
  \includegraphics[width=\sizefigggg\linewidth]{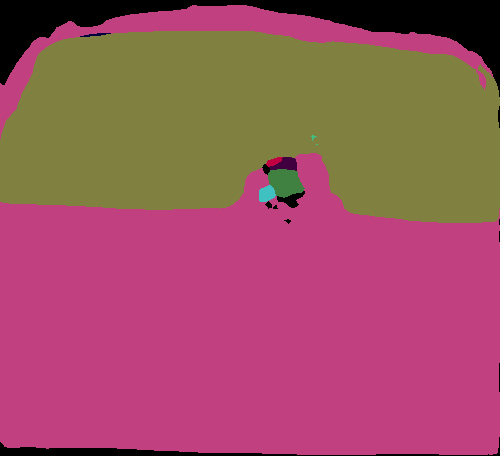} \\
  
    \includegraphics[width=\sizefigggg\linewidth]{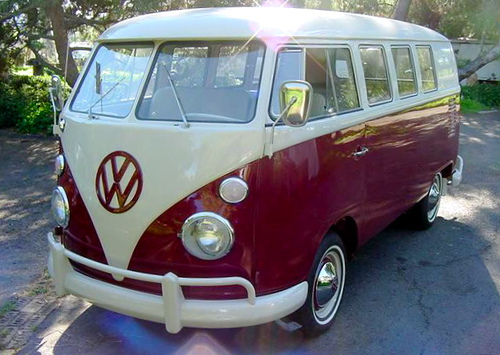} &
  \includegraphics[width=\sizefigggg\linewidth]{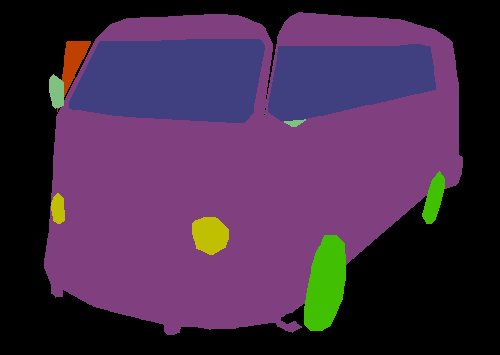} &
  \includegraphics[width=\sizefigggg\linewidth]{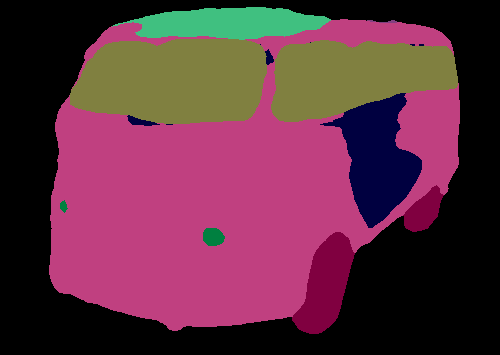} &
  \includegraphics[width=\sizefigggg\linewidth]{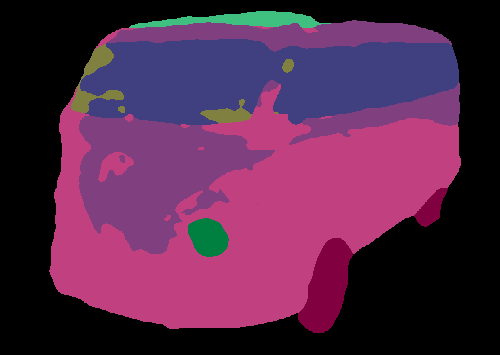} &
  \includegraphics[width=\sizefigggg\linewidth]{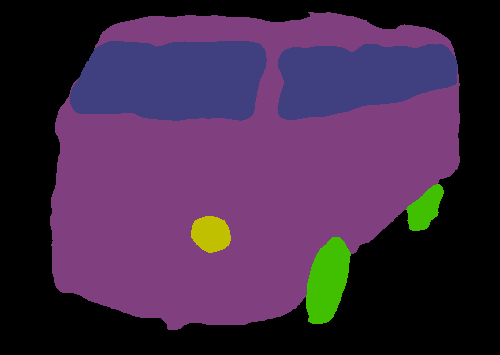} \\
  
    \includegraphics[width=\sizefigggg\linewidth]{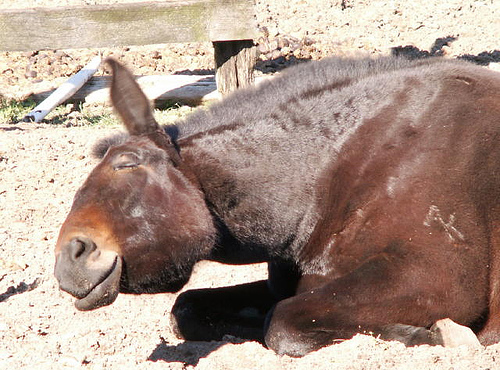} &
  \includegraphics[width=\sizefigggg\linewidth]{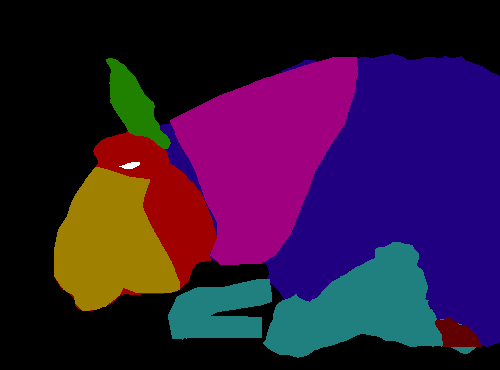} &
  \includegraphics[width=\sizefigggg\linewidth]{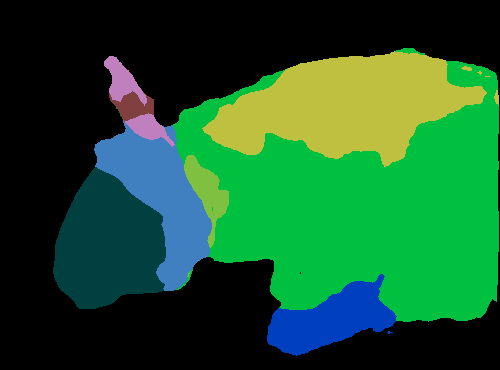} &
  \includegraphics[width=\sizefigggg\linewidth]{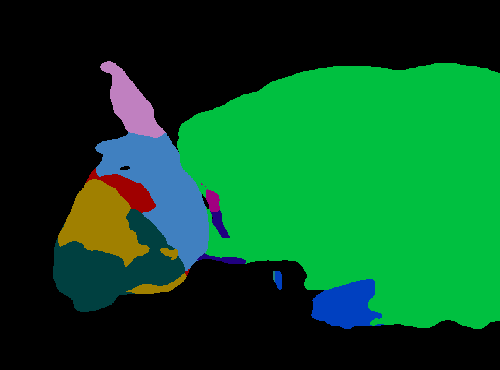} &
  \includegraphics[width=\sizefigggg\linewidth]{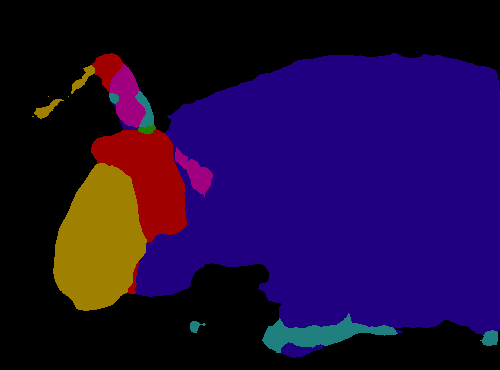} \\
  
      \includegraphics[width=\sizefigggg\linewidth]{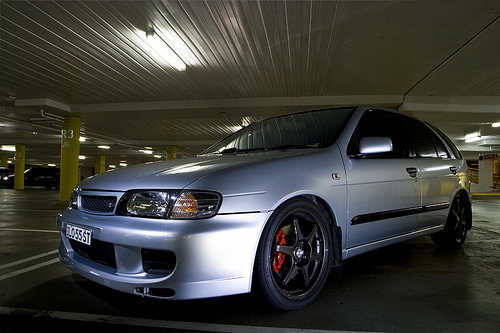} &
  \includegraphics[width=\sizefigggg\linewidth]{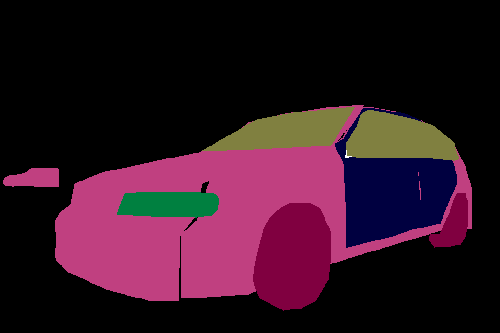} &
  \includegraphics[width=\sizefigggg\linewidth]{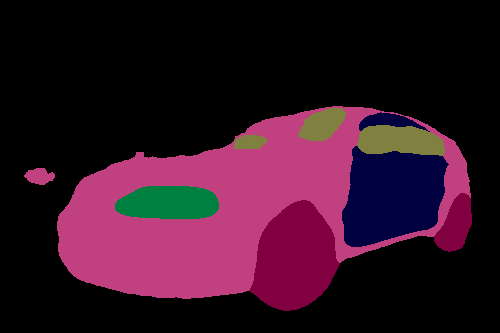} &
  \includegraphics[width=\sizefigggg\linewidth]{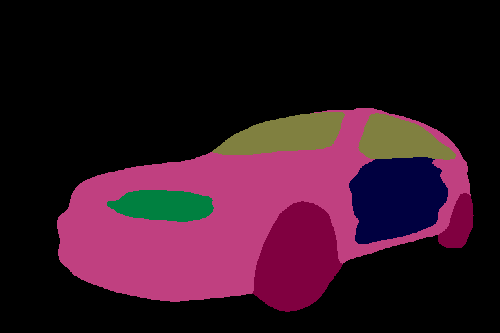} &
  \includegraphics[width=\sizefigggg\linewidth]{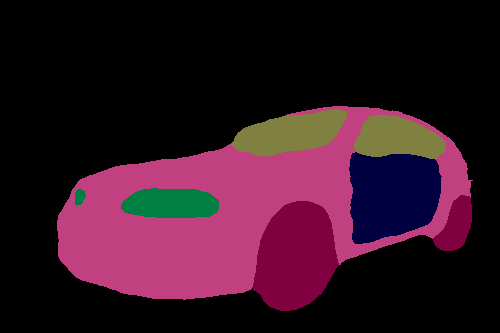} \\
  
  \includegraphics[width=\sizefigggg\linewidth]{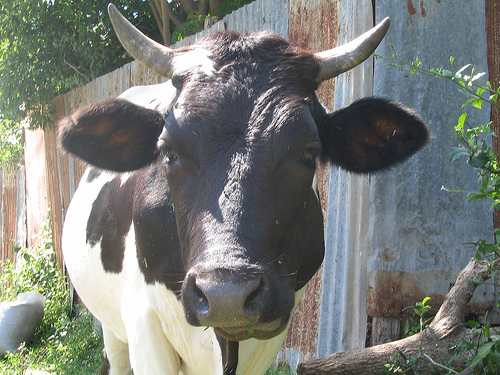} &
  \includegraphics[width=\sizefigggg\linewidth]{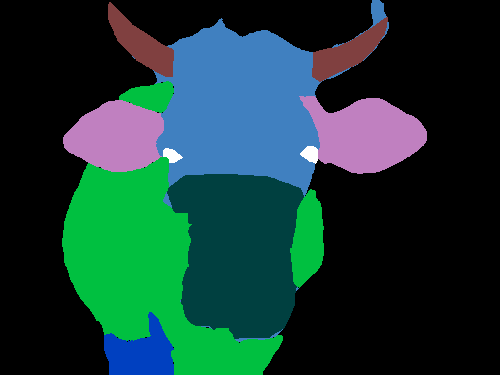} &
  \includegraphics[width=\sizefigggg\linewidth]{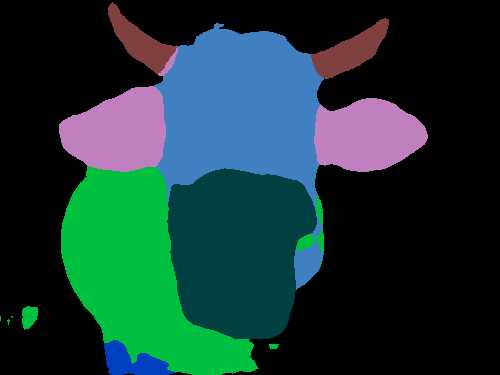} &
  \includegraphics[width=\sizefigggg\linewidth]{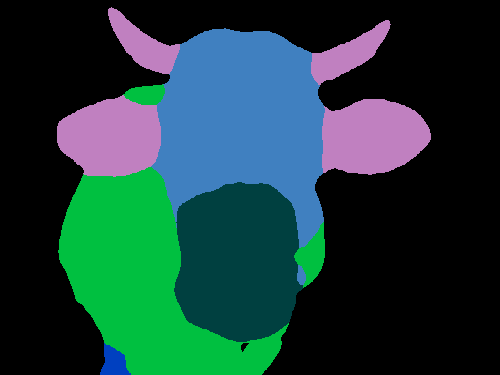} &
  \includegraphics[width=\sizefigggg\linewidth]{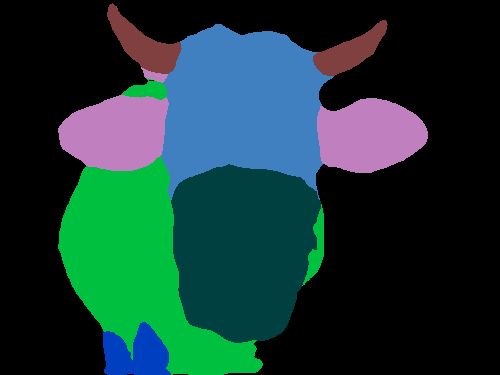} \\

    \includegraphics[width=\sizefigggg\linewidth]{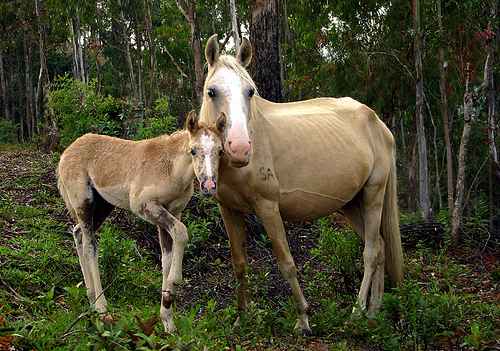} &
  \includegraphics[width=\sizefigggg\linewidth]{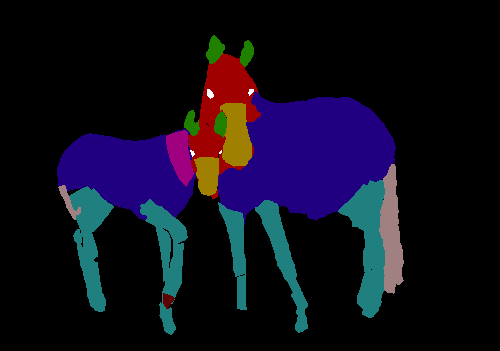} &
  \includegraphics[width=\sizefigggg\linewidth]{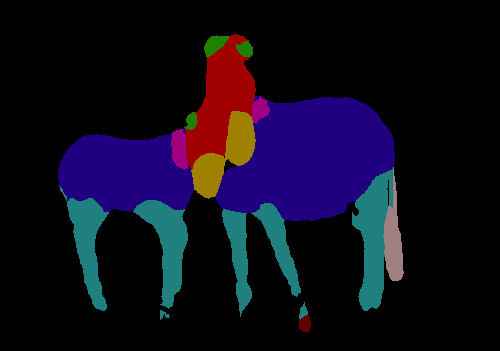} &
  \includegraphics[width=\sizefigggg\linewidth]{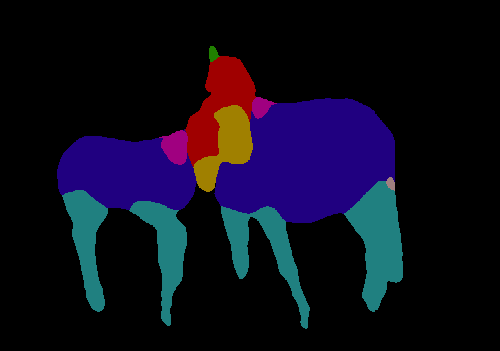} &
  \includegraphics[width=\sizefigggg\linewidth]{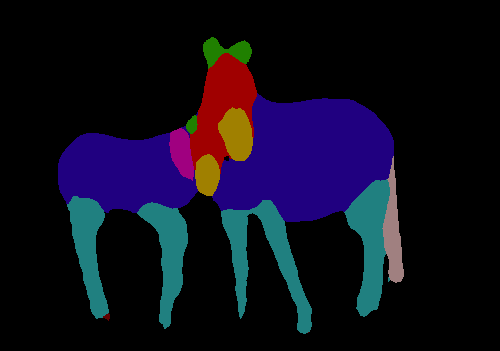} \\
  
  \includegraphics[width=\sizefigggg\linewidth]{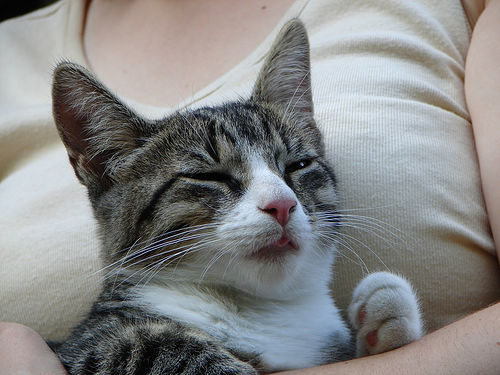} &
  \includegraphics[width=\sizefigggg\linewidth]{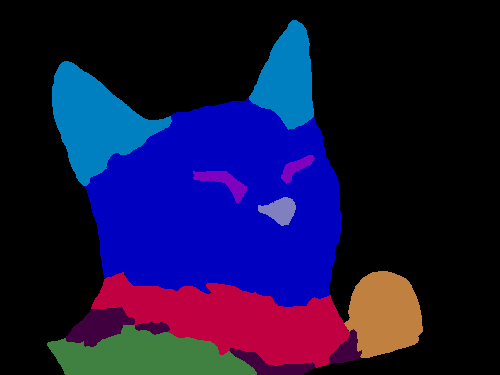} &
  \includegraphics[width=\sizefigggg\linewidth]{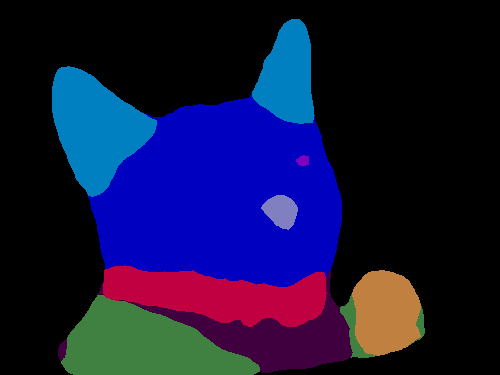} &
  \includegraphics[width=\sizefigggg\linewidth]{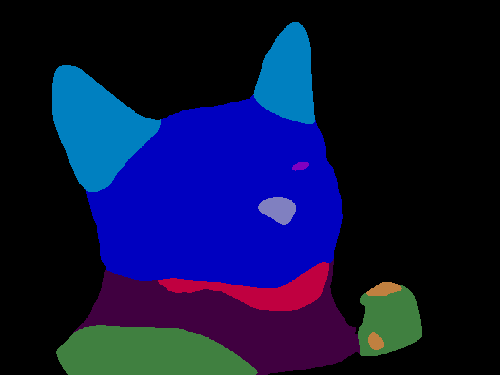} &
  \includegraphics[width=\sizefigggg\linewidth]{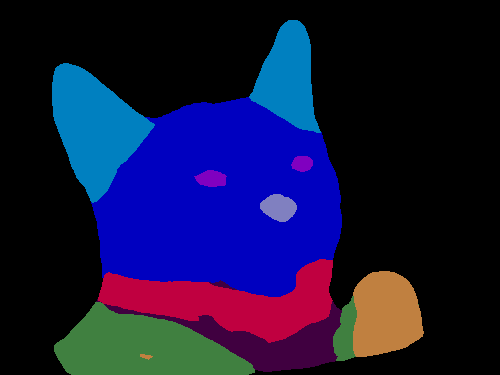} \\

  \includegraphics[width=\sizefigggg\linewidth]{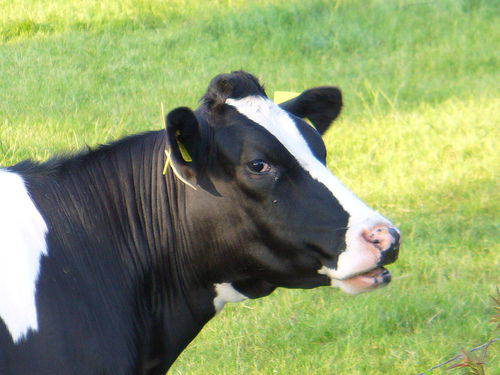} &
  \includegraphics[width=\sizefigggg\linewidth]{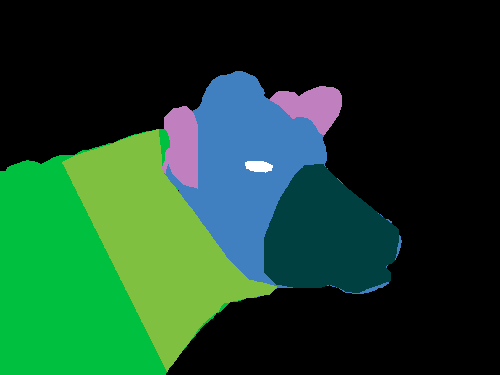} &
  \includegraphics[width=\sizefigggg\linewidth]{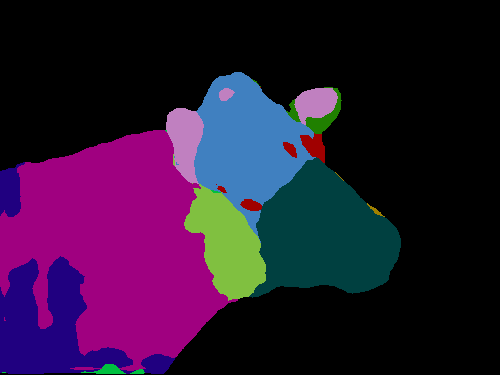} &
  \includegraphics[width=\sizefigggg\linewidth]{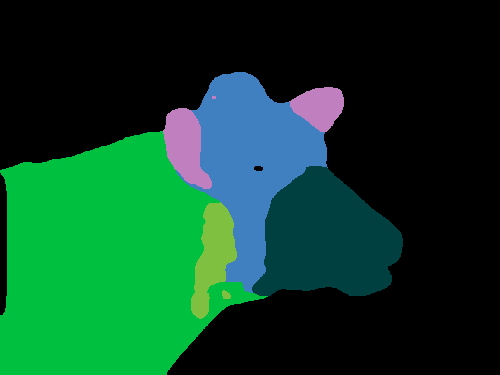} &
  \includegraphics[width=\sizefigggg\linewidth]{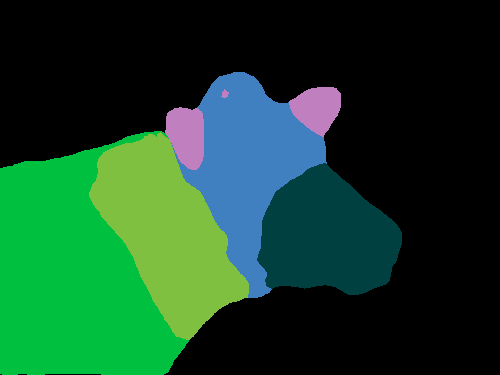} \\

 \end{tabular}
\vspace{-0.1cm}
\caption{Qualitative results on sample scenes on the Pascal-Part-108 dataset (\textit{best viewed in colors}).}
\label{fig_suppl:Pascal_part_108}
\end{figure}

\begin{table}[htbp]
\caption{Comparison in terms of mIoU, mCA and mPA on Pascal-Part-108.}
\label{tab_suppl:pascal_part_108_mean}
\setlength{\tabcolsep}{6pt}
\centering
\begin{tabular}{|l|c|c|c|}
\hline
Method & mIoU & mPA & mCA \\\hline
Baseline \cite{chen2017rethinking} & $41.36$ & $88.57$ & $50.51$ \\
BSANet \cite{zhao2019multi} & $42.95$ & $89.52$ & $51.71$\\
GMNet & $\mathbf{45.80}$ & $\mathbf{90.32}$ & $\mathbf{55.68}$\\\hline
\end{tabular}
\end{table}

Then, in Figure~\ref{fig_suppl:Pascal_part_108} we report some additional qualitative results. The effect of the object-level semantic embedding network is particularly evident in the first $4$ rows. In row 1, a challenging image is presented where both the baseline and BSANet are not able to correctly identify the table. In rows $2$ and $3$, GMNet generates cleaner segmentation maps exploiting object-level priors which help to disambiguate between cars and buses. In row 4, the baseline and BSANet predict cat's parts in spite of horse's parts which are partially identified by our method. 

The graph matching module is much more effective on this dataset because contains many small-sized parts. We can verify this from the sixth to the last row. In row $6$, the \textit{cow\_horns} and \textit{cow\_body} are badly localized and labelled both by the baseline and by BSANet. However, the graph matching component on the reciprocal spatial relationship between these parts and the others guides the network to properly localize and label such parts. In row $7$ our framework is able to well localize horse's parts and especially the challenging \textit{horse\_tail} part. In the second-last row, GMNet correctly identifies difficult cat's parts such as \textit{cat\_eyes} and \textit{cat\_paws} thanks to the graph matching module. In the last row, the semantic embedding module allows our method to identify the cow and, at the same time, the graph matching module allows to correctly localize the spatial relations among all the parts.